\documentclass[10pt,twocolumn,letterpaper]{article}

\usepackage{cvpr}              

\usepackage[dvipsnames]{xcolor}

\definecolor{cvprblue}{rgb}{0.21,0.49,0.74}
\usepackage[pagebackref,breaklinks,colorlinks,citecolor=cvprblue]{hyperref}
\usepackage{graphicx}
\usepackage{float}
\usepackage[font=footnotesize,labelfont=bf]{caption}
\usepackage{amssymb}
\usepackage{pifont}
\newcommand{\xmark}{\ding{55}}%

\def\paperID{***} 
\def\confName{CVPR}
\def\confYear{2024}

\definecolor{custom_blue}{HTML}{3369E8}
\definecolor{custom_yellow}{HTML}{EEB211}
\definecolor{custom_green}{HTML}{009925} 
\definecolor{custom_red}{HTML}{FF5765}

\begin{document}
\title{The Manga Whisperer: Automatically Generating Transcriptions for Comics}

\author{
  Ragav Sachdeva\quad Andrew Zisserman\vspace{5pt}\\
  {\normalsize Visual Geometry Group, Dept.\ of Engineering Science, University of Oxford}
}


\twocolumn[{
    \renewcommand\twocolumn[1][]{#1}
    \maketitle
    \centering
    \vspace{-10pt}
    \includegraphics[width=\textwidth]{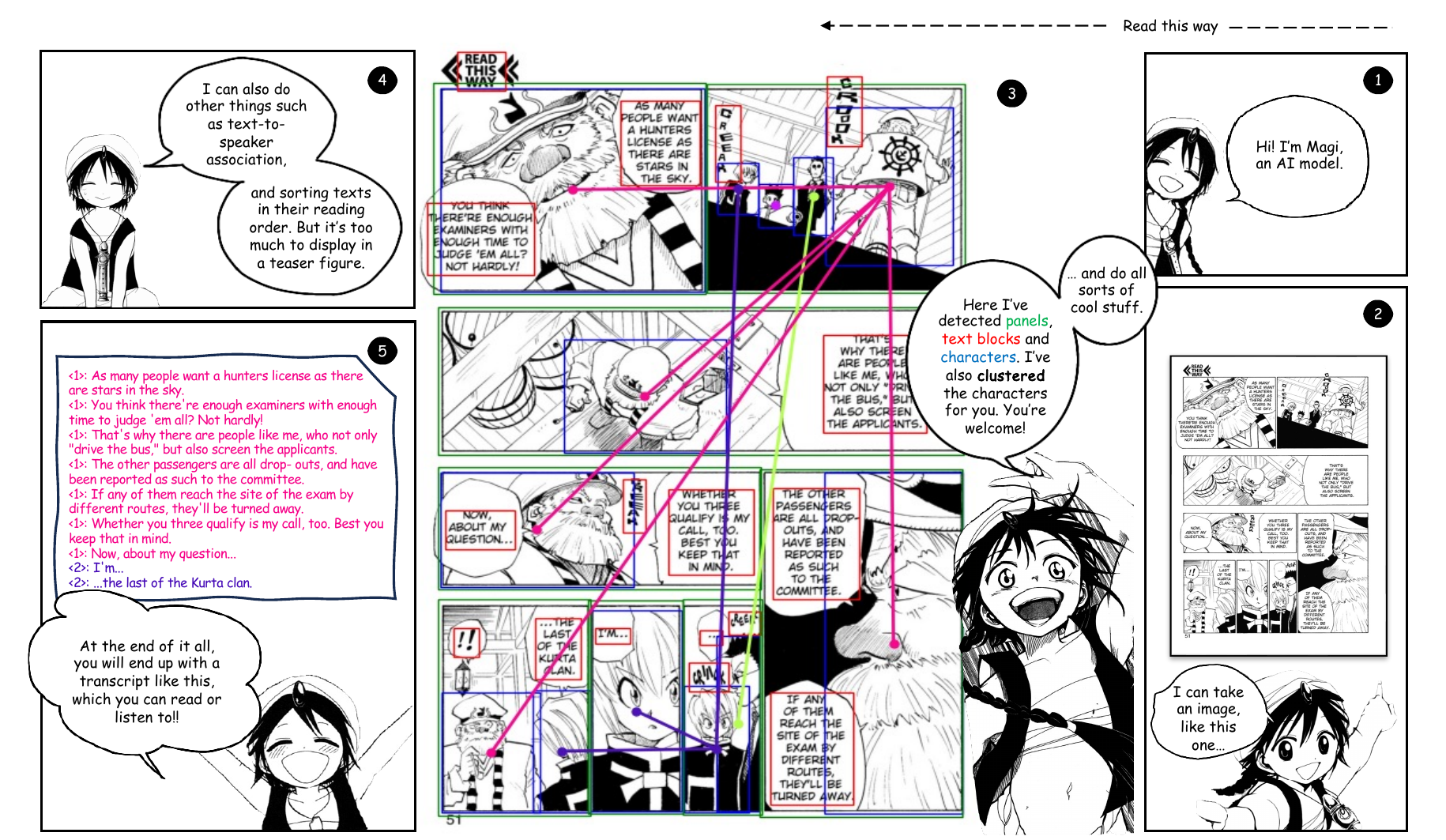}
    \captionof{figure}{Given a manga page, our model is able to: (a) detect panels, text blocks and character boxes; (b) cluster character boxes by their identity; (c) associate texts to their speaker; and (d) generate a dialogue transcription in the correct reading order. Here we show the predicted panels \textcolor{custom_green}{(in green)}, text blocks \textcolor{custom_red}{(in red)} and characters \textcolor{custom_blue}{(in blue)} on a page from \textit{Hunter $\times$ Hunter} by Yoshihiro Togashi. The predicted character identity associations are shown by lines joining the character box centres. For visual clarity, we do not explicitly show the text to speaker associations but provide the generated transcript.}
    \label{fig:teaser}
    \vspace{20pt}
    }
    ]

\begin{abstract}
In the past few decades, Japanese comics, commonly referred to as Manga, have transcended both cultural and linguistic boundaries to become a true worldwide sensation. Yet, the inherent reliance on visual cues and illustration within manga renders it largely inaccessible to individuals with visual impairments. In this work, we seek to address this substantial barrier, with the aim of ensuring that manga can be appreciated and actively engaged by everyone. Specifically, we tackle the problem of diarisation i.e.\ generating a transcription of who said what and when, in a fully automatic way.

To this end, we make the following contributions: (1) we present a unified model, \textbf{Magi}, that is able to (a) detect panels, text boxes and character boxes, (b) cluster characters by identity (without knowing the number of clusters apriori), and (c) associate dialogues to their speakers; (2) we propose a novel approach that is able to sort the detected text boxes in their reading order and generate a dialogue transcript; (3) we annotate an evaluation benchmark for this task using publicly available [English] manga pages. The code, evaluation datasets and the pre-trained model can be found at: \url{https://github.com/ragavsachdeva/magi}.
\end{abstract} 

\section{Introduction}
From billboards in New York's Times Square to murals in Paris, manga characters, in all their colourful splendor, are everywhere. It is undeniably evident that manga's ubiquity extends far and wide, transcending cultural and geographic boundaries to become a cherished and adored art form, captivating the hearts of enthusiasts from diverse walks of life and backgrounds across the globe. Manga's popularity can be attributed to its rich and diverse content that covers an extensive range of genres and themes, catering to a broad and varied audience. 
From action-packed \textit{shounen} manga to the intricate storytelling of \textit{seinen} manga, 
the depth and breadth of storytelling in manga is unparalleled. These diverse set of themes allow readers to connect with a manga on a personal level, making it a universal form of entertainment. There is truly a manga for everyone... except for people with visual impairments (PVI).

\noindent The interest of PVI to be able to access comics is well documented~\cite{life_philipp_meyer,textured_images,thermoforming,star_wars_audiobook,ramaprasad2023comics}.
In a recent study~\cite{accesscomics}, conducted to understand the accessibility issues that PVI experience with comics, when the participants were asked to select the most important piece of information they wished to know while reading a comic, the majority responded with scene descriptions, followed by transcriptions, facial expressions of characters, etc. With the recent advances in computer vision and deep learning the time is right to attempt to extract this information automatically, and this work is a step towards realising that goal.

\noindent The task of ``understanding" manga fully automatically, to then describe it to PVI, is {\em very} challenging. This involves solving a series of problems including panel detection, panel ordering, text detection, OCR, character detection, character identification, text-to-speaker association, scene captioning, action recognition, etc.
As shown in Figure~\ref{fig:teaser}, this is extremely challenging. Characters are often drawn with varying fidelity and can be seen from different views (front, back, profile etc.), in various poses, and possibly only partially due to occlusions from speech balloons or artistic choices. It is also quite common to have non-human characters (e.g.\ monsters). Furthermore, text blocks may or may not be inside speech balloons which, in turn, may or may not have tails indicating the speaker. The artistic layout, special visual effects and resolution pose additional difficulties.
As humans, we are able to use context and deductive reasoning to understand these complex manga, but for machines it is still very challenging.

\noindent In this work, our objective is {\em diarisation} -- to be able to generate a transcription, page by page, of who said what in the veridical order, to convey the story on that page. To this end, we develop a model, \textit{Magi}, for this task using a unified architecture. In order to achieve diarisation, we must address a major portion of the challenges noted above.  Specifically, we tackle the problem of detection (panel, text and characters) and association (character-character and character-text) and treat it as a graph generation problem (where detections are the ``nodes" and their associations are the ``edges"). This completes the \textit{computer vision} aspect of the description. We then mobilise prior knowledge of manga layout to generate a transcription from the graph with the correct ordering.
This separation into two stages (graph generation, and transcription) enables a lighter model to be used than a single model trained end-to-end for the task.

\noindent To summarise, we make the following contributions: (1) we present a novel model that ingests a manga page and is able to (a) detect panels, characters and text blocks, (b) cluster characters by their identity (without making any assumptions about the number of unique identities), and (c) associate dialogues to their speakers; (2) we propose a novel method to generate a dialogue transcript using the extracted panels, text blocks and associations; (3) we create a challenging evaluation benchmark, called \textit{PopManga}, comprising of pages from 80+ popular manga by various artists known for their complexity and detailed story-telling, and demonstrate the superior performance of our method over prior works.

\begin{figure*}[ht]
    \centering
    \includegraphics[width=\linewidth]{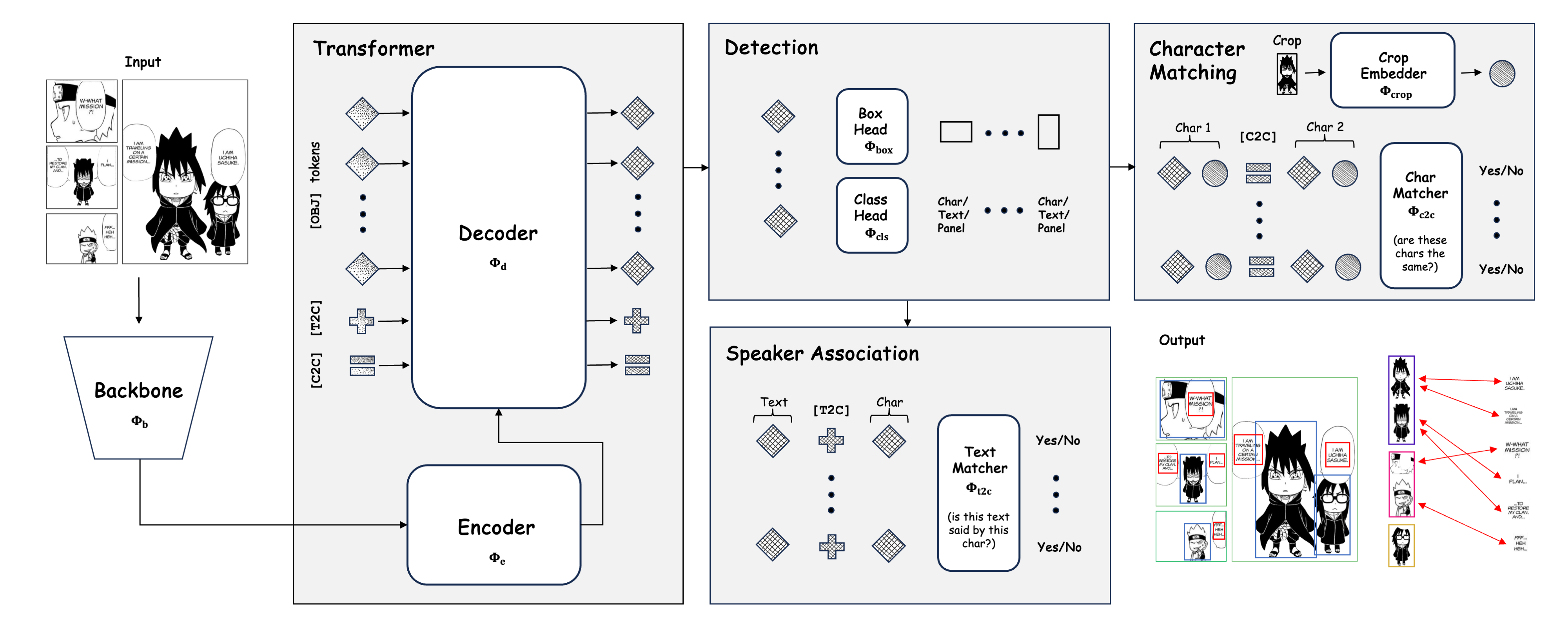}
    \caption{\textbf{The \textit{Magi} Architecture:} Given a manga page as input, our model predicts bounding boxes for panels, text blocks and characters, and associates the detected character-character and text-character pairs. The model ingests a high resolution manga page as input to a CNN backbone, followed by a transformer encoder-decoder resulting in $N\times$\texttt{[OBJ]} + \texttt{[C2C]} + \texttt{[T2C]} tokens. The \texttt{[OBJ]} tokens are processed by the detection heads (box and class) to obtain the bounding boxes and their classifications. The \texttt{[OBJ]} tokens corresponding to detected objects are then processed in pairs, along with \texttt{[C2C]} and \texttt{[T2C]}, by a character matching module and a speaker association module respectively resulting in character clusters and diarisation.}
    \label{fig:model_arch}
\end{figure*}

\section{Related Works}

The problem of comic understanding (not limited to manga) is not new.  \textbf{\textit{(i)}} There are several existing works that propose solutions for panel detection~\cite{pang2014robust, ogawa2018object, he2018end, wang2015comic, nguyen2019comic, rigaud2015speech}, text/speech balloon detection~\cite{nguyen2019comic, piriyothinkul2019detecting, ogawa2018object, manga109coo} and character detection~\cite{topal2022domain, ogawa2018object, inoue2018cross, jiang2022decoupled}. Specifically for manga text and panel detection, the most successful approaches train their models on the Manga109~\cite{manga109} dataset. This dataset consists of $21000+$ images and, in its most recent version, provides annotations for panels, text blocks, characters (face and body), character identification, text to speaker associations~\cite{manga109dialog}, onomatopoeia~\cite{manga109coo} etc. For character detection, the state-of-the-art method~\cite{topal2022domain} leverages domain adaptation and transfer learning techniques to achieve impressive results.
\textbf{\textit{(ii)}} For character re-identification, \cite{tsubota2018adaptation} proposes an unsupervised approach to cluster manga character faces. \cite{qin2019progressive} introduces a method that leverages transfer learning and iteratively refines labels for positives and negatives to train their model. 
\cite{zhang2022unsupervised} propose an unsupervised face-body clustering method that incorporates the panel index of the characters for spatial-temporal information. Recently, \cite{soykan2023identity} use a SimCLR~\cite{chen2020simple} style approach to train a unified model that works for both face and body character crops. \textbf{\textit{(iii)}} For speaker identification, \cite{rigaud2015speech} construct a bipartite graph of characters and texts and treat speaker association as a matching problem based on the Euclidean distance. Recently, \cite{manga109dialog} released dialogue annotations for Manga109 and propose a scene-graph based approach to address the speaker identification problem. We compare our method with prior works that are available publicly.

\section{Detection and Association}
\label{sec:graph}
Given a manga page, our goal is to produce a transcript of who said what and when in a fully automatic way. Therefore, as a requirement, the model must be aware of the various components that constitute a manga page---particularly the panels, characters and text blocks---and how they are related. To this end, we need to \textit{detect} the panels, characters and text blocks (i.e.\ where they are on the page), as well as \textit{associate} them -- character-character association (i.e.\ clustering), and text-character association (i.e.\ speaker identification). The \textit{detection} part is relatively straightforward; there is a plethora of methods that can be used~\cite{meng2021conditional, carion2020end, zhu2020deformable, zhou2019objects}. The tricky part is the \textit{association}, and particularly character-character association.\newline

\noindent\textbf{Architecture overview:} We formulate these tasks as graph generation problem and propose a unified model, \textit{Magi}, that is able to simultaneously detect panels, text blocks and characters (nodes of the graph), and perform character-character matching and text-character matching (edges of the graph). The architecture is illustrated in Figure~\ref{fig:model_arch}.  The input is an entire high resolution manga page, that is processed by a CNN-based backbone to obtain a spatial feature map. These dense feature descriptors are then processed by a DETR~\cite{carion2020end} style encoder-decoder transformer, resulting in $N$~\texttt{[OBJ]} tokens and $2$ special tokens---\texttt{[C2C]}, \texttt{[T2C]}---that we introduce, inspired by~\cite{shit2022relationformer}. The \texttt{[OBJ]} tokens are processed by the detection heads (box and class) to obtain the bounding boxes and their classifications (into character, text, panel, or background). The \texttt{[OBJ]} tokens corresponding to detected objects are then processed in pairs, along with \texttt{[C2C]} and \texttt{[T2C]}, by a character matching module and a speaker association module respectively, which essentially answer the question \textit{``Is there an edge between a given pair of \texttt{[OBJ]} nodes?"}, resulting in character clusters and speaker associations.\newline

\noindent\textbf{Design choices:} We perform ``in-context" detection and association, where the network ingests the entire page. In contrast,  
previous methods for character re-identification~\cite{qin2019progressive, zhang2022unsupervised, soykan2023identity, tsubota2018adaptation} operate on {\em crops} of characters (either faces or full-body) and use metric learning to learn desirable feature representations which can then be used to compute a similarity score given two character crops. This approach is limited in two ways: (a) metric learning based solutions are good for the retrieval task where there is query image and a large gallery of candidates to retrieve matches from, but they are not well suited to make hard decision boundaries (i.e.\ form clusters) especially when the number of clusters is unknown apriori and the number of data points per cluster is too few (as is the case for a single manga page); (b) only operating on character crops loses the surrounding visual cues on the page that can assist with grouping characters. In Sec.~\ref{sec:results} we compare our model with crop-based methods and validate our design choices. 

\subsection{Architecture details}

\noindent\textbf{Backbone} is a Convolutional Neural Network (CNN), represented by $\Phi_{b}(\cdot)$, that ingests a high resolution input manga page $I \in \mathbb{R}^{3\times H\times W}$ and extracts dense feature descriptors  $f \in \mathbb{R}^{c \times hw}$.\newline

\noindent\textbf{Transformer Encoder} is a DETR-style transformer encoder, represented by $\Phi_{e}(\cdot)$, which processes content embeddings $f$ from the backbone with the aim of improving them, resulting in features $g \in \mathbb{R}^{c \times hw}$.\newline

\noindent\textbf{Transformer Decoder} is a DETR-style transformer decoder with conditional cross-attention~\cite{meng2021conditional}, represented by $\Phi_{d}(\cdot)$. It processes $N+2$ \textit{query} tokens as input, which consist of $N\times$\texttt{[OBJ]} + \texttt{[C2C]} + \texttt{[T2C]} tokens, that undergo a series of self-attention layers that perform interactions between the query latents, and cross-attention layers that attend to the contextualised image features $g$. The idea here is that the \texttt{[OBJ]} tokens learn to attend to specific spatial positions in the image, encoding the information pertaining to that region, while \texttt{[C2C]} and \texttt{[T2C]} learn to encode the interactions between these objects in a pooled fashion. The output of $\Phi_{d}(\cdot)$ is represented by $h^{obj} \in \mathbb{R}^{c \times N}, h^{c2c} \in \mathbb{R}^{c \times 1}, h^{t2c} \in \mathbb{R}^{c \times 1}$.\newline

\noindent\textbf{Detection Head} is a two-component module that processes $h^{obj}$ using (i) a MLP $\Phi_{box}(\cdot)$ to regress the object locations, and (ii) a linear layer $\Phi_{cls}(\cdot)$ to classify the objects.\newline

\noindent\textbf{Crop-Embedding Module} is a ViT~\cite{dosovitskiy2020image} model, represented by $\Phi_{crop}(\cdot)$ that, given a crop of the detected character $i$, outputs an embedding vector $c_i\in \mathbb{R}^{c \times 1}$.\newline

\noindent\textbf{Character Matching Module} is an MLP, represented by $\Phi_{c2c}$, followed by sigmoid activation that outputs a score $A_{ij}^{char}$ of character $i$ being the same as character $j$. It does so by processing a single feature vector that is obtained by concatenating $h_i^{obj}, c_i, h^{c2c}, h_j^{obj}, c_j$,
where $c_i, c_j$ are the crop embedding vectors for the character $i, j$ respectively (crop is obtained using the predicted boxes). 
The final binary prediction of whether $i$ and $j$ are the same characters is obtained by thresholding $A_{ij}^{char}$ using a hyperparamter $\tau$.\newline

\noindent\textbf{Speaker Association Module} is also a MLP, represented by $\Phi_{t2c}$, followed by sigmoid activation that outputs a score $A_{ij}^{text}$ of text $i$ being said by character $j$. It does so by processing a single feature vector that is obtained by concatenating $h_i^{obj}, h^{t2c}, h_j^{obj}$. The final speaker prediction for text $i$ is given by
$$\text{Speaker}_i = \arg \max_{j} A_{i}^{text}$$

\subsection{Training}
\label{sec:training}
Achieving end-to-end training for this model, i.e.\ to jointly optimise the model for each of the detection and association tasks, requires a dataset that contain bounding boxes for panels, text blocks, characters, as well as annotations for character clusters and text-speaker associations, for each page. In Sec.~\ref{sec:datasets} we describe how we obtain two datasets with these annotations: (1) {\em Mangadex-1.5M}, a large-scale (1.5M images) dataset with (possibly noisy) pseudo-annotations generated automatically; and 
(2) {\em PopManga}, a smaller dataset with 55000+ images in the dev set, and high quality human annotations.
 We train the model in two steps, first using Mangadex-1.5M and then using PopManga \textit{Dev} subset, in a supervised fashion.\newline

\noindent\textbf{Implementation:} The input image is resized such that the shorter side is $800$px and the longer side is resized appropriately to preserve the aspect ratio but capped at $1333$px. It is then projected to a spatial feature map by the backbone $\Phi_{b}(\cdot)$, which has the ResNet-50~\cite{he2016deep}  architecture. The dimensions of the spatial feature map depend on the resolution of the input image (e.g.\ for a resized $1100 \times 800$px image, the spatial map is $35\times25$).
The spatial dimension of this feature map is flattened and the channel dimension is projected to $256$ before passing it to the transformer. The transformer encoder-decoder  $\Phi_{e}(\cdot),\Phi_{d}(\cdot)$ have 6 layers each, with the hidden dimension of $256$, and $8$ attention heads. The number of \texttt{[OBJ]} decoder queries is $N=300$.
$\Phi_{b}(\cdot),\Phi_{e}(\cdot),\Phi_{d}(\cdot),\Phi_{box}(\cdot),\Phi_{cls}(\cdot)$, are initialised using Conditional-DETR~\cite{meng2021conditional} weights. The crop-embedding module $\Phi_{crop}(\cdot)$ has 12 layers, with the hidden dimension of 768 and 12 attention heads, and is initialised using MAE~\cite{he2021masked} encoder weights. The input to $\Phi_{crop}(\cdot)$ is resized to $224\times 224$. The character matching and speaker association modules $\Phi_{c2c}, \Phi_{t2c}$ are both 3-layered MLPs initialised randomly.
We set the character matching threshold $\tau = 0.65$. Our training objective for detection is the same as in~\cite{meng2021conditional}. We further apply Binary Cross Entropy loss to the outputs of our character matching and speaker association heads. Additionally, we apply Supervised Contrastive Loss~\cite{khosla2020supervised} to the per-page embeddings from the crop-embedding module.
We trained our model end-to-end, on $2\times$A40 GPUs using AdamW~\cite{loshchilov2017decoupled} optimizer with both learning rate and weight decay of 0.0001, and effective batch size of 16.

\section{Transcript Generation}
\label{sec:transcript}
Once the bounding boxes for panels, characters and text boxes have been extracted, along with the character clusters and speaker associations, generating a transcript from them is really just an OCR + Sorting problem. To sort the text boxes into their reading order, we leverage prior knowledge about manga layout. Specifically, manga pages are read from top to bottom, \textit{right to left} (note that the reading order is different from western comics which are read top to bottom, left to right). Given this, we order the text boxes in two steps: (a) order the panels to give the relative ordering of text boxes belonging to different panels, (b) order the text boxes within each panel. After ordering the text boxes, we perform OCR to extract the content of the texts and finally generate the transcript using all the computed data.\newline

\noindent\textbf{Panel Ordering:} Previous methods~\cite{ordering, hinami2021towards} use ``cuts" to recursively split the panels into horizontal and vertical partitions. The idea is to construct a tree by recursively splitting the panels using (1) a horizontal line, (2) a vertical line, in that order of priority, and then traverse the tree in a specific order (top panels before bottom for horizontal cuts, right panels before left for vertical cuts) to get the reading order. While this method works in most cases, it fails in the case of overlapping panels, where there is no clean way to ``cut" the page. We propose an improved algorithm to sort the panels in their reading order that overcomes this issue. We very briefly describe our approach below and provide more details in the appendix.

\begin{figure}[H]
    \centering
    \includegraphics[width=\columnwidth]{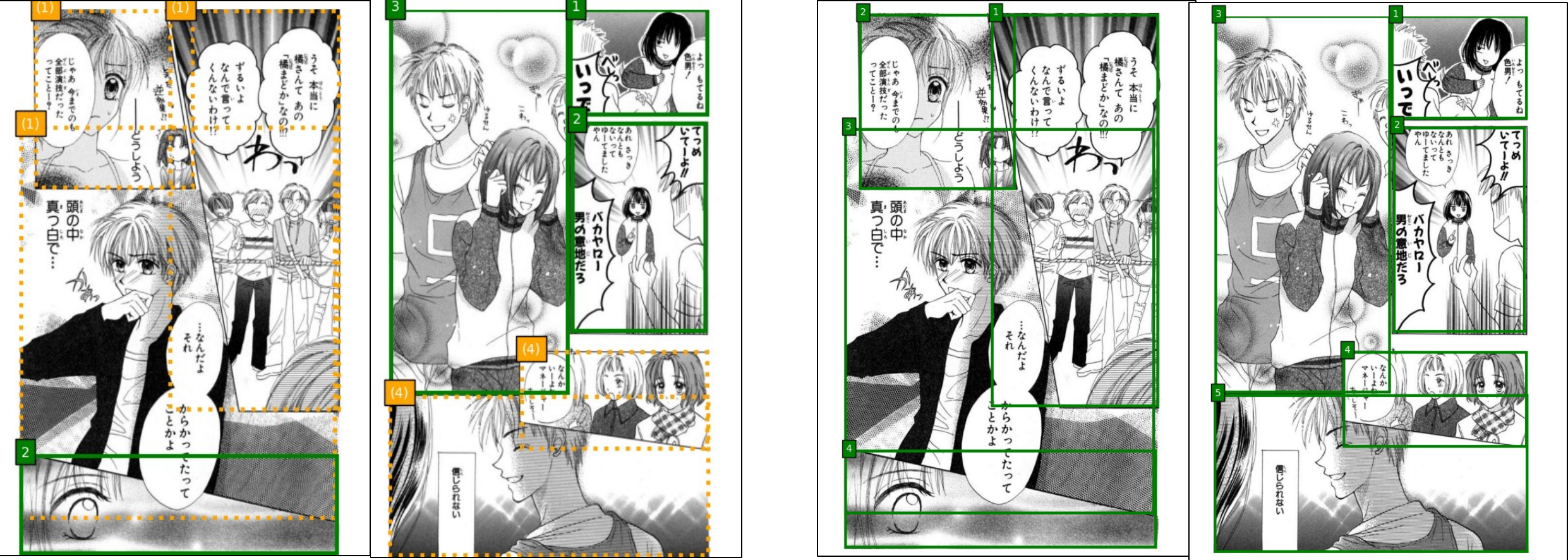}
    \caption{\textbf{Panel ordering:} On the left are the ordering predictions using~\cite{ordering} and on the right using ours. Images: Prism Heart \copyright Mai Asatsuki.}
    \label{fig:ordering}
\end{figure}

\textit{Approach:} The idea behind our approach is to represent the panels as a directed acyclic graph (DAG) where each directed edge represents the \textit{relative} reading order between the connecting panels, and then apply Topological Sorting~\cite{kahn1962topological} to this DAG. The reason our method works more robustly than an entirely ``cut" based method is because it can infer the relative order of two intersecting panels by relaxing the constraint of ``strictly" above/below/left/right to ``largely" above/below/left/right. By considering panels in pairs, we do not require \textit{global} clean ``cuts" to exist. In Figure~\ref{fig:ordering}, we show the ordering prediction using our method vs prior works.

\noindent\textbf{Intra-Panel Text Ordering:} There is no absolute rule that determines the order of texts within the panel. As a general rule of thumb, however, the texts within the panel are read in the order of their distance from the top-right corner of the panel. Following previous works~\cite{ordering, hinami2021towards}, we use this heuristic to sort texts within each panel.\newline

\noindent\textbf{OCR:} While performing OCR is not the primary goal of this work, we train an OCR model for completeness. Specifically, we finetune the TrOCR~\cite{li2021trocr} model on custom synthetic data generated using the pipeline described in~\cite{manga_ocr}. We provide more details in the appendix.

\begin{figure*}
    \centering
    \includegraphics[width=\linewidth]{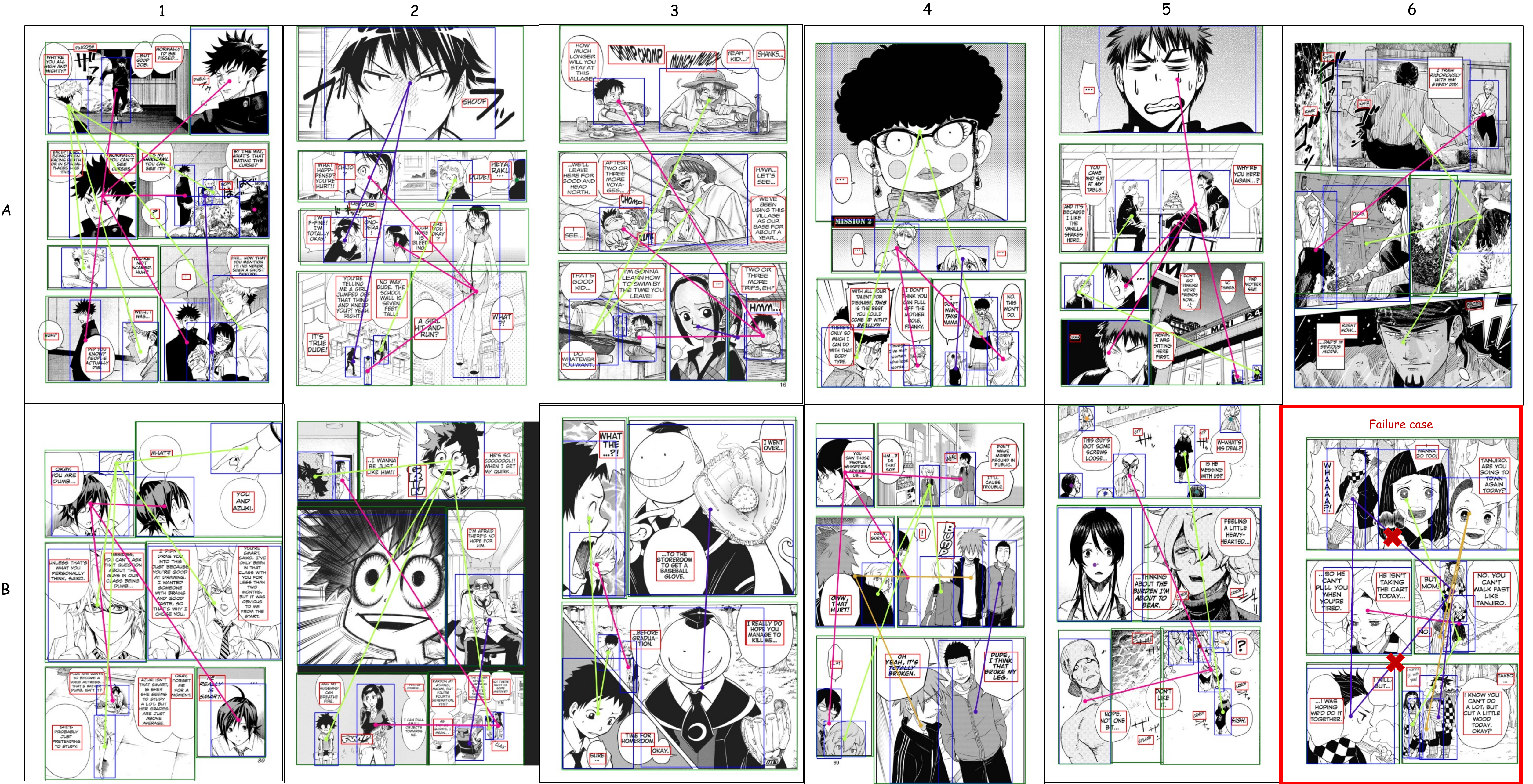}
    \caption{Bounding box predictions determined  by the \textit{Magi} model for \textcolor{custom_blue}{characters}, \textcolor{custom_red}{text blocks} and \textcolor{custom_green}{panels}, as well as clustering predictions (as nodes and edges). For the purposes of visualisation, we remove redundant connections for characters that are already connected via transitivity. Best viewed digitally. Notice that the model has successfully matched characters despite occlusion/partial visibility (girl: A4, hand: B1), changing viewpoints (boy: A5, girl: A2), and varying fidelity (girl: A1, boy: B2). The model can also detect non-human characters (dog-like creatures: A1, octopus-like creature: B3). We also show a failure case where the model incorrectly matches two different characters (one is wearing checked scarf, while the other is wearing checked coat, they are brothers and look similar).}
    \label{fig:predictions}
\end{figure*}

\begin{figure*}
    \centering
    \includegraphics[width=\linewidth]{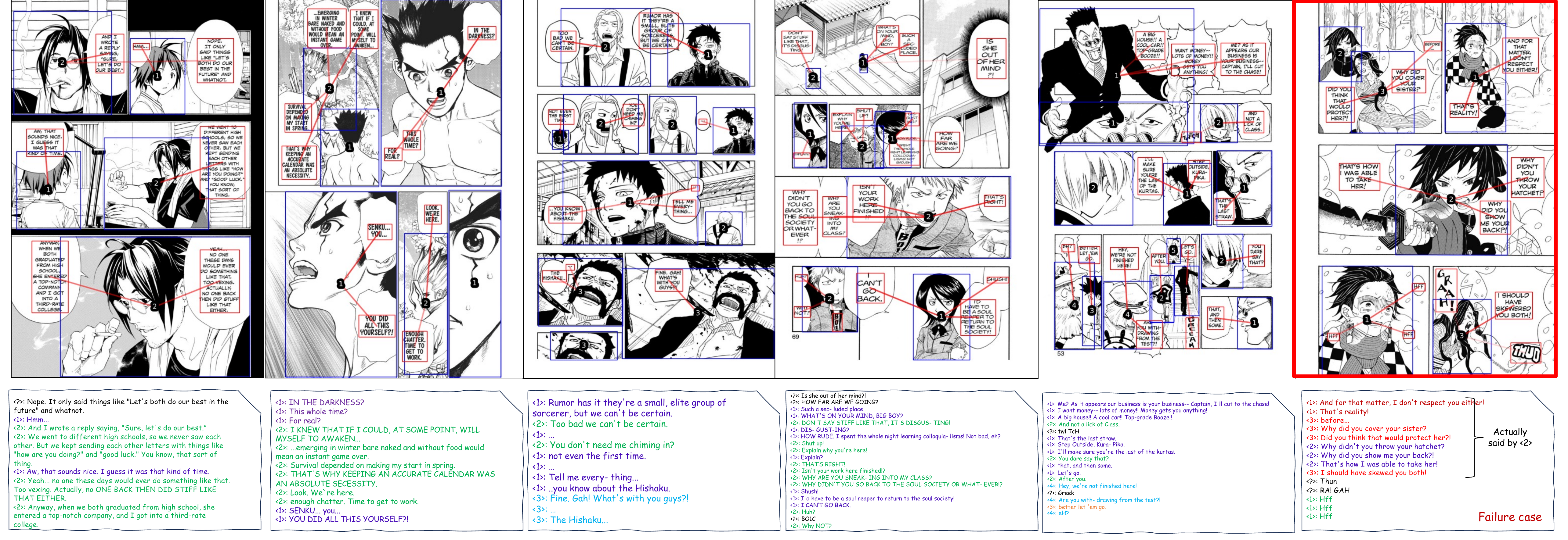}
    \caption{Text to speaker predictions generated by the \textit{Magi}  model. Each predicted text box is connected to a predicted character box using a line. The opacity of the line reflects the confidence of the model (the darker the line, the more confident the model is). Each predicted character box has a number at its centre based on the clustering predictions. We also show the final generated transcript. Note that all the dialogues are in the correct reading order. For text to speaker predictions that have a low confidence score ($<0.4$) we replace the predicted speaker with $\langle?\rangle$ in the generated transcript and let the reader infer it from context. Best viewed digitally.}
    \label{fig:predictions_diarisation}
\end{figure*}

\section{Datasets}
\label{sec:datasets}

For training and evaluation we require
datasets that contain bounding boxes for panels, text blocks, characters, as well as annotations for character clusters and text-speaker associations, for each page. In the following we describe three datasets that cover these requirements: an existing dataset {\em Manga109}~\cite{manga109}, and two new datasets that we introduce, {\em PopManga} and {\em Mangadex-1.5M}.
We provide an overview of these datasets in Sec.~\ref{sec:dataset_overview}, and how we collected and annotated the new datasets in Sec.~\ref{sec:curation_and_annotation}. 

\subsection{Dataset overview}
\label{sec:dataset_overview}
The three datasets are summarised in Table~\ref{tab:datasets}.\newline

\begin{table}[H]
\centering
\resizebox{\columnwidth}{!}{%
\begin{tabular}{cc|cccccc}
\hline
\multicolumn{2}{c|}{Dataset}                      & \multicolumn{6}{c}{Annotations}                                                                                  \\ \hline
\multicolumn{1}{c|}{name}              & \#images & \multicolumn{1}{c|}{source} & \#P            & \#T     & \#C     & \#C2C                 & \#T2C                 \\ \hline
\multicolumn{1}{c|}{PopManga (Dev)}  & 55,393   & \multicolumn{1}{c|}{Human}  & $264,067^\dag$ & 459,615 & 487,367 & 397,054               & $52,011^\ddag$        \\
\multicolumn{1}{c|}{PopManga (Test-S)}    & 1,136    & \multicolumn{1}{c|}{Human}  & \xmark         & 11,843  & 11,498  & 9,960                 & 7,537                 \\
\multicolumn{1}{c|}{PopManga (Test-U)}   & 789      & \multicolumn{1}{c|}{Human}  & \xmark         & 9,000   & 7,280   & 5,819                 & 6,090                 \\ \hline
\multicolumn{1}{c|}{Manga109~\cite{manga109}} & 21,204    & \multicolumn{1}{c|}{Human}  & 104,566          & 147,894     & 157,512  & 167,020                & 131,039                   \\ \hline
\multicolumn{1}{c|}{Mangadex-1.5M}     & 1.57M    & \multicolumn{1}{c|}{Pseudo} & 6.6M           & 12M     & 14M     & ${}^{\dagger\dagger}$ & ${}^{\dagger\dagger}$ \\ \hline
\end{tabular}%
}
\caption{\textbf{Training and evaluation datasets.}
In the columns we provide the number of annotations for each task P = Panels, T = Texts, C = Characters, C2C = Character-character positive pairs, T2C = Text-character pairs. ${}^{\dag}$ automatically generated. ${}^{\ddag}$ for 8,488 images only. ${}^{\dagger\dagger}$ mined using heuristics dynamically.}
\label{tab:datasets}
\end{table}

\noindent\textbf{PopManga:} A new dataset of 57000+ manga pages from $80+$ different series that are considered to be in the list the top manga of all time~\cite{mal_top_manga}. These manga are typically known for their complex storytelling and intricate art style, thus making them very challenging.
We split PopManga into 3 subsets. The \textit{Dev} subset is used for training and validation, and the remaining two subsets---\textit{Test-S} (test seen), \textit{Test-U} (test unseen)---are used for testing such that \textit{Test-S} consists of 30 chapters from 15 series, where \textit{other} chapters from these series are seen during training, while \textit{Test-U} consists of 20 chapters from 10 series that the model never sees during training.

We intend the \textit{Test-S} and \textit{Test-U} images to serve as a public benchmark for evaluation. 
For this reason,  they are from chapters that are available freely, publicly and officially on Manga Plus by Shueisha~\cite{MangaPlus}, thus enabling future academic research and comparisons.
A comprehensive list of manga name and chapters in \textit{Dev}, \textit{Test-S}, \textit{Test-U} is available in the appendix.\newline

\noindent\textbf{Manga109~\cite{manga109}:} An existing dataset of $21000+$ images across $109$ manga volumes, that provides all the detection and association annotations we require. However, in this work, we are mainly interested in the diarisation of English versions of manga which are not available in Manga109. Therefore, we only use Manga109 for some evaluation purposes.\newline

\noindent\textbf{Mangadex-1.5M:} A large scale dataset of $\approx 1.5$ million manga pages spanning multiple genres and art styles. A large majority of these are not made by professionals, and therefore, on average the images are ``simpler" than PopManga. We use this dataset for pretraining.

\subsection{Dataset curation and annotations}
\label{sec:curation_and_annotation}

\noindent\textbf{PopManga}\newline

\noindent\textbf{Curation:} 57000+ images are downloaded from various sources including, but not limited to, Mangadex~\cite{MangaDex} and MangaPlus~\cite{MangaPlus}.\newline

\noindent\textbf{Annotation:} We get human annotators to label text boxes, character boxes, character-character association and text-character association using a modified version of the LISA annotation tool~\cite{Dutta19a}. For panels, we first finetune DETR~\cite{carion2020end} to predict panels using Manga109 and then use it to automatically generate pseudo panel annotations for the \textit{Dev} subset of PopManga (we found these generated panels to be of satisfactory quality, thus reducing annotation cost). In the appendix we include more information on the annotation process.\newline

\noindent\textbf{Mangadex-1.5M}\newline

\noindent\textbf{Curation:} The $\approx 1.5$ million \textit{unlabelled} manga pages are downloaded from Mangadex~\cite{MangaDex} using their official API. During curation, we only query for manga which are available in English and have a safe-for-work content rating.\newline

\noindent\textbf{Annotation:} For the collected images we do not have ground truth annotations. Instead, we generate and mine pseudo-annotations for them. \textit{(1) Detection:} We prompt Grounding-DINO~\cite{grounding_dino} (which has a strong zero-shot open-vocabulary detection performance) to detect  bounding boxes for characters (using ``human" and ``character" as prompt) and text blocks (using ``text" as prompt). We use a finetuned DETR to generated psuedo panel annotations, as done for PopManga \textit{Dev}. \textit{(2) Character-Character association}: We mine positive/negative character pairs for each page using the following strategy: (a) if character A and character B are in the same panel, they are not the same character (negatives); (b) if character A and character B are nearest neighbours of each other in the model's latent space, then A and B are the same characters (positives), as long as A and B are not in the same panel; (c) if $A=B$ and $B\neq C$, then $A\neq C$ (more negatives via transitivity). \textit{(3) Text-Character association}: Here we simply match each text block to the closest character. This is not always correct, but it is correct more often than not.\newline

\section{Results}
\label{sec:results}
In the following we report results on the set of tasks that are needed to realise our goal of diarisation. For baselines we use state-of-the-art models that are available publicly.

\subsection{Test sets and Baselines}
\noindent \textbf{Character detection}\newline
\noindent\textit{Datasets:} PopManga (Test-S), PopManga (Test-U) and Manga109 (same test split as~\cite{topal2022domain}).\newline
\noindent \textit{Baselines:} We compare our model against state-of-the-art character detection model DASS~\cite{topal2022domain}, and zero-shot results from Grounding-DINO~\cite{grounding_dino}.\newline

\noindent \textbf{Text detection}\newline
\noindent \textit{Datasets:} PopManga (Test-S) and PopManga (Test-U). We do not report results on Manga109 as it does not contain English texts.\newline
\noindent \textit{Baselines:} We compare our model against the zero-shot results from Grounding-DINO~\cite{grounding_dino}.\newline

\noindent \textbf{Panel detection}\newline
\noindent \textit{Datasets:} Manga109 (same test split as~\cite{topal2022domain}).
\newline
\noindent \textit{Baselines:} We compare our model against the zero-shot results from Grounding-DINO~\cite{grounding_dino}.\newline

\noindent \textbf{Character clustering}\newline
\noindent \textit{Datasets:} PopManga (Test-S), PopManga (Test-U) and Manga109 (same test split as~\cite{tsubota2018adaptation}).\newline
\noindent \textit{Baselines:} We compare our model against AMFR~\cite{tsubota2018adaptation}, which is a manga face clustering model, ZACI~\cite{zaci20}, which is a zero-shot anime face identification model. In addition, we use large pretrained vision encoders, CLIP~\cite{radford2021learning} and DINO-V2~\cite{oquab2023dinov2}, to evaluate their feature representation for manga characters. Note that the baselines operate on cropped characters, unlike our method which operates directly on the entire manga page.\newline

\noindent \textbf{Speaker association}\newline
\noindent \textit{Datasets:} PopManga (Test-S) and PopManga (Test-U). We do not report results on Manga109 as it does not contain English texts.\newline
\noindent \textit{Baselines:} To the best of our knowledge, there are no openly available models that can be adapted for our purposes. Instead we use ``match each text box to the closest character" heuristic as a simple baseline.

\subsection{Evaluation Metrics}

\noindent\textbf{Detection:} We follow the object detection literature and evaluate our method using the average precision metric, as defined by~\cite{lin2014microsoft}. For a predicted box to be positive, we use an IOU threshold of 0.5. Since methods being evaluated generate a different number of bounding boxes, we use the top 100 predictions only for a fair comparison.

\noindent\textbf{Character clustering:} We evaluate our method using several metrics that are commonly used in the metric learning literature. These metrics can be categorised into (1) retrieval based metrics---Mean Reciprocal Rank (MRR), Mean Average Precision at R (MAP@R), Precision at 1 (P@1), R-Precision (R-P)---that essentially operate on nearest neighbours of a query sample, (2) clustering based metrics---Adjusted Mutual Info (AMI), Normalised Mutual Info (NMI)---that evaluate the quality of predicted clusters and hard decision boundaries. We compute these metrics for each manga page and average over the entire test set.

\noindent\textbf{Speaker association:} We evaluate our method using Recall@\#text metric, as formulated in~\cite{manga109dialog}. This is similar to the Recall@$K$ metric~\cite{lu2016visual}, except $K$ is the number of texts per page.  This metric is computed for each manga page and averaged over the entire test set.

\begin{table}[H]
\centering
\resizebox{\columnwidth}{!}{%
\begin{tabular}{ccc|cc|cc}
                                    & \multicolumn{2}{c|}{PopManga (Test-S)}  & \multicolumn{2}{c|}{PopManga (Test-U)} & \multicolumn{2}{c}{Manga109}      \\ \hline
\multicolumn{1}{c|}{method}         & \multicolumn{1}{c|}{Char}   & Text   & \multicolumn{1}{c|}{Char}   & Text   & \multicolumn{1}{c|}{Body} & Panel \\ \hline
\multicolumn{1}{c|}{DASS~\cite{topal2022domain}}           & \multicolumn{1}{c|}{0.8410} & -      & \multicolumn{1}{c|}{0.8580} & -      & \multicolumn{1}{c|}{\textbf{0.9251}}     &   -  \\
\multicolumn{1}{c|}{Grounding-DINO~\cite{grounding_dino}} & \multicolumn{1}{c|}{0.7250}       &   0.7922  & \multicolumn{1}{c|}{0.7420}     &   0.8301     & \multicolumn{1}{c|}{0.7985}      & 0.5131     \\
\multicolumn{1}{c|}{Magi [Ours]}    & \multicolumn{1}{c|}{\textbf{0.8485}} & \textbf{0.9227} & \multicolumn{1}{c|}{\textbf{0.8615}} & \textbf{0.9208} & \multicolumn{1}{c|}{0.9015}      & \textbf{0.9357}
\end{tabular}%
}
\caption{\textbf{Detection Results.} We report the average precision results, which have an upper bound of 1.0.}
\label{tab:detection}
\end{table}

\begin{table}[H]
\centering
\resizebox{\columnwidth}{!}{%
\begin{tabular}{ccccccc}
\hline
method &
  AMI &
  NMI &
  MRR &
  MAP@R &
  P@1 &
  R-P \\ \hline
\multicolumn{1}{c|}{} &
  \multicolumn{6}{c}{PopManga (Test-S)} \\ \hline
\multicolumn{1}{c|}{DINO-V2~\cite{oquab2023dinov2}} &
  \multicolumn{1}{c|}{0.0704} &
  \multicolumn{1}{c|}{0.6219} &
  \multicolumn{1}{c|}{0.6549} &
  \multicolumn{1}{c|}{0.3870} &
  \multicolumn{1}{c|}{0.4851} &
  0.4390 \\
\multicolumn{1}{c|}{CLIP~\cite{radford2021learning}} &
  \multicolumn{1}{c|}{0.1053} &
  \multicolumn{1}{c|}{0.5405} &
  \multicolumn{1}{c|}{0.7237} &
  \multicolumn{1}{c|}{0.4757} &
  \multicolumn{1}{c|}{0.5836} &
  0.5189 \\
\multicolumn{1}{c|}{ZACI~\cite{zaci20}} &
  \multicolumn{1}{c|}{0.0510} &
  \multicolumn{1}{c|}{0.5656} &
  \multicolumn{1}{c|}{0.6470} &
  \multicolumn{1}{c|}{0.3806} &
  \multicolumn{1}{c|}{0.4687} &
  0.4354 \\
\multicolumn{1}{c|}{AMFR~\cite{tsubota2018adaptation}} &
  \multicolumn{1}{c|}{0.1550} &
  \multicolumn{1}{c|}{0.5355} &
  \multicolumn{1}{c|}{0.7404} &
  \multicolumn{1}{c|}{0.5004} &
  \multicolumn{1}{c|}{0.6085} &
  0.5397 \\
\multicolumn{1}{c|}{Magi (crop only) [Ours]} &
  \multicolumn{1}{c|}{0.4892} &
  \multicolumn{1}{c|}{0.7178} &
  \multicolumn{1}{c|}{0.9008} &
  \multicolumn{1}{c|}{0.7840} &
  \multicolumn{1}{c|}{0.8423} &
  0.8008 \\
\multicolumn{1}{c|}{Magi [Ours]} &
  \multicolumn{1}{c|}{\textbf{0.6574}} &
  \multicolumn{1}{c|}{\textbf{0.8501}} &
  \multicolumn{1}{c|}{\textbf{0.9312}} &
  \multicolumn{1}{c|}{\textbf{0.8439}} &
  \multicolumn{1}{c|}{\textbf{0.8884}} &
  \textbf{0.8555} \\ \hline
\multicolumn{1}{c|}{} &
  \multicolumn{6}{c}{PopManga (Test-U)} \\ \hline
\multicolumn{1}{c|}{DINO-V2~\cite{oquab2023dinov2}} &
  \multicolumn{1}{c|}{0.0806} &
  \multicolumn{1}{c|}{0.6885} &
  \multicolumn{1}{c|}{0.6281} &
  \multicolumn{1}{c|}{0.3621} &
  \multicolumn{1}{c|}{0.4420} &
  0.4135 \\
\multicolumn{1}{c|}{CLIP~\cite{radford2021learning}} &
  \multicolumn{1}{c|}{0.1057} &
  \multicolumn{1}{c|}{0.5691} &
  \multicolumn{1}{c|}{0.7145} &
  \multicolumn{1}{c|}{0.4623} &
  \multicolumn{1}{c|}{0.5636} &
  0.5027 \\
\multicolumn{1}{c|}{ZACI~\cite{zaci20}} &
  \multicolumn{1}{c|}{0.0767} &
  \multicolumn{1}{c|}{0.3354} &
  \multicolumn{1}{c|}{0.6409} &
  \multicolumn{1}{c|}{0.3797} &
  \multicolumn{1}{c|}{0.4603} &
  0.4276 \\
\multicolumn{1}{c|}{AMFR~\cite{tsubota2018adaptation}} &
  \multicolumn{1}{c|}{0.1828} &
  \multicolumn{1}{c|}{0.6141} &
  \multicolumn{1}{c|}{0.7570} &
  \multicolumn{1}{c|}{0.5319} &
  \multicolumn{1}{c|}{0.6235} &
  0.5676 \\
\multicolumn{1}{c|}{Magi (crop only) [Ours]} &
  \multicolumn{1}{c|}{0.4862} &
  \multicolumn{1}{c|}{0.7326} &
  \multicolumn{1}{c|}{0.9061} &
  \multicolumn{1}{c|}{0.7926} &
  \multicolumn{1}{c|}{0.8477} &
  0.8076 \\
\multicolumn{1}{c|}{Magi [Ours]} &
  \multicolumn{1}{c|}{\textbf{0.6527}} &
  \multicolumn{1}{c|}{\textbf{0.8503}} &
  \multicolumn{1}{c|}{\textbf{0.9347}} &
  \multicolumn{1}{c|}{\textbf{0.8557}} &
  \multicolumn{1}{c|}{\textbf{0.8936}} &
  \textbf{0.8656} \\ \hline
\multicolumn{1}{c|}{} &
  \multicolumn{6}{c}{Manga109 (face)} \\ \hline
\multicolumn{1}{c|}{DINO-V2~\cite{oquab2023dinov2}} &
  \multicolumn{1}{c|}{0.1788} &
  \multicolumn{1}{c|}{0.5795} &
  \multicolumn{1}{c|}{0.7451} &
  \multicolumn{1}{c|}{0.5108} &
  \multicolumn{1}{c|}{0.5918} &
  0.5468 \\
\multicolumn{1}{c|}{CLIP~\cite{radford2021learning}} &
  \multicolumn{1}{c|}{0.1990} &
  \multicolumn{1}{c|}{0.6502} &
  \multicolumn{1}{c|}{0.7698} &
  \multicolumn{1}{c|}{0.5466} &
  \multicolumn{1}{c|}{0.6263} &
  0.5798 \\
\multicolumn{1}{c|}{ZACI~\cite{zaci20}} &
  \multicolumn{1}{c|}{0.2428} &
  \multicolumn{1}{c|}{0.5231} &
  \multicolumn{1}{c|}{0.7970} &
  \multicolumn{1}{c|}{0.5951} &
  \multicolumn{1}{c|}{0.6679} &
  0.6245 \\
\multicolumn{1}{c|}{AMFR~\cite{tsubota2018adaptation}} &
  \multicolumn{1}{c|}{0.2814} &
  \multicolumn{1}{c|}{0.5739} &
  \multicolumn{1}{c|}{0.8167} &
  \multicolumn{1}{c|}{0.6272} &
  \multicolumn{1}{c|}{0.7003} &
  0.6536 \\
\multicolumn{1}{c|}{Magi (crop only) [Ours]} &
  \multicolumn{1}{c|}{0.3429} &
  \multicolumn{1}{c|}{0.6078} &
  \multicolumn{1}{c|}{0.8381} &
  \multicolumn{1}{c|}{0.6653} &
  \multicolumn{1}{c|}{0.7341} &
  0.6896 \\
\multicolumn{1}{c|}{Magi [Ours]} &
  \multicolumn{1}{c|}{\textbf{0.6254}} &
  \multicolumn{1}{c|}{\textbf{0.7843}} &
  \multicolumn{1}{c|}{\textbf{0.9148}} &
  \multicolumn{1}{c|}{\textbf{0.8151}} &
  \multicolumn{1}{c|}{\textbf{0.8558}} &
  \textbf{0.8279} \\ \hline
\multicolumn{1}{c|}{} &
  \multicolumn{6}{c}{Manga109 (body)} \\ \hline
\multicolumn{1}{c|}{DINO-V2~\cite{oquab2023dinov2}} &
  \multicolumn{1}{c|}{0.1077} &
  \multicolumn{1}{c|}{0.5248} &
  \multicolumn{1}{c|}{0.6963} &
  \multicolumn{1}{c|}{0.4337} &
  \multicolumn{1}{c|}{0.5183} &
  0.4814 \\
\multicolumn{1}{c|}{CLIP~\cite{radford2021learning}} &
  \multicolumn{1}{c|}{0.1643} &
  \multicolumn{1}{c|}{0.5034} &
  \multicolumn{1}{c|}{0.7542} &
  \multicolumn{1}{c|}{0.5164} &
  \multicolumn{1}{c|}{0.6027} &
  0.5591 \\
\multicolumn{1}{c|}{ZACI~\cite{zaci20}} &
  \multicolumn{1}{c|}{0.1090} &
  \multicolumn{1}{c|}{0.4504} &
  \multicolumn{1}{c|}{0.6919} &
  \multicolumn{1}{c|}{0.4289} &
  \multicolumn{1}{c|}{0.5116} &
  0.4767 \\
\multicolumn{1}{c|}{AMFR~\cite{tsubota2018adaptation}} &
  \multicolumn{1}{c|}{0.2622} &
  \multicolumn{1}{c|}{0.6054} &
  \multicolumn{1}{c|}{0.8076} &
  \multicolumn{1}{c|}{0.5980} &
  \multicolumn{1}{c|}{0.6845} &
  0.6293 \\
\multicolumn{1}{c|}{Magi (crop only) [Ours]} &
  \multicolumn{1}{c|}{0.5690} &
  \multicolumn{1}{c|}{0.7694} &
  \multicolumn{1}{c|}{0.9237} &
  \multicolumn{1}{c|}{0.8259} &
  \multicolumn{1}{c|}{0.8721} &
  0.8389 \\
\multicolumn{1}{c|}{Magi [Ours]} &
  \multicolumn{1}{c|}{\textbf{0.6345}} &
  \multicolumn{1}{c|}{\textbf{0.8202}} &
  \multicolumn{1}{c|}{\textbf{0.9383}} &
  \multicolumn{1}{c|}{\textbf{0.8567}} &
  \multicolumn{1}{c|}{\textbf{0.8966}} &
  \textbf{0.8667} \\ \hline
\end{tabular}%
}
\caption{\textbf{Character Clustering Results.} We report results using several metrics. They all have an upper bound of 1.0.}
\label{tab:char_clus}
\end{table}

\begin{table}[H]
\centering
\resizebox{0.8\columnwidth}{!}{%
\begin{tabular}{l|l|l}
method            & PopManga (Test-S) & PopManga (Test-U) \\ \hline
shortest distance &       0.7758         &        0.7659         \\ \hline
Magi [Ours]       &          \textbf{0.8448}      &              \textbf{0.8313}  
\end{tabular}%
}
\caption{\textbf{Speaker Association Results.} We report the Recall@\#text results, which have an upper bound of 1.0.}
\label{tab:diarisation}
\end{table}

\subsection{Performance and Discussion}
We report quantitative results in Tables~\ref{tab:detection},~\ref{tab:char_clus}, and~\ref{tab:diarisation} and show qualitative results in Figures~\ref{fig:predictions} and~\ref{fig:predictions_diarisation}.
These results demonstrate the efficacy of our model and its impressive performance over the baselines.

\noindent \textit{Detection:} For the task of detecting texts and panels, our method performs markedly better compared to existing solutions. For characters, our model’s performance is comparable to DASS~\cite{topal2022domain}, surpassing it in 2 out of 3 test sets, with a slight concession in the remaining third. Notably, Grounding-DINO~\cite{grounding_dino} shows a strong zero-shot detection performance for characters and texts. However, its effectiveness takes a discernible dip when applied to panels. 

\noindent \textit{Character clustering:} The biggest strength of our model, \textit{Magi}, lies in its ability to perform ``in-context" character clustering, establishing itself as a state-of-the-art solution across all metrics and test sets. Particularly the metric of interest is AMI, which evaluates clustering performance and the model's ability to form decisive boundaries. Note that AMI is adjusted such that a random clustering has a score of 0 whereas in NMI random clusters may still get a positive score. Our model's AMI results significantly outshine those of existing methods, showcasing superiority by several magnitudes. 
Additionally, the results of our crop-only embedding module, though superior to baselines, fall short of the performance achieved by our holistic model, \textit{Magi}, with context.
This underscores our hypothesis that operating only on crops loses visual cues on the page that can assist in clustering. 

In Figure~\ref{fig:predictions}, our model demonstrates its prowess in character clustering, successfully handling scenarios where characters are observed from disparate viewpoints, with partial visibility, at various scales, and diverse expressions. Notably, a failure case is presented where the model erroneously matched two characters with similar checked clothing, emphasising the challenges posed by subtle visual details, even when the characters are distinct (in this case, brothers). Nevertheless, it is evident that the model is able to pick up on multiple cues for character matching, including clothing, hair style, and the faces.

\noindent \textit{Speaker Association:} In Figure~\ref{fig:predictions_diarisation}, we show that our method can adeptly match texts to their respective speakers. We also demonstrate the use of the confidence values to filter cases where the model is unsure, which includes non-dialogue texts (e.g.\ sound effects) or texts with no character in the panel. The task of matching texts to their speakers in manga is challenging, often necessitating an understanding of conversation history and context to disambiguate speakers. 

\section{Conclusion}
In this study, our primary objective was to improve the accessibility of manga for individuals with visual impairments. Tackling the complex task of diarisation, we have laid the groundwork for a fully automated transcription of manga content, enabling active engagement for everyone, irrespective of their visual abilities. 

Looking ahead, numerous promising avenues for future research beckon. Particularly, we anticipate leveraging the language understanding capabilities of Large Language Models (LLMs) to enhance diarisation by incorporating conversation history and context. 

\newpage
\noindent \textbf{Acknowledgements:} We would like to thank Cindy Seuk and Anhad Sachdeva for their assistance with annotation quality assurance, and Gyungin (Noel) Shin, Sindhu Hegde, and Paul Engstler for proof-reading the paper. This research is supported by EPSRC Programme Grant VisualAI EP/T028572/1 and a Royal Society Research Professorship RP\textbackslash R1\textbackslash191132.

{
    \small
    \bibliographystyle{ieeenat_fullname}
    \bibliography{main}
}

\clearpage
\onecolumn
\setcounter{page}{1}
\maketitlesupplementary

\appendix

\section{Outline}
\label{sec:outline}
\begin{itemize}
    \item Section~\ref{supp_sec:panel_ordering} describes the algorithm used to order panels in more detail.
    \item Section~\ref{supp_sec:popmanga} catalogues the list of manga chapters in \textit{Dev}, \textit{Test-S} and \textit{Test-U} subsets of PopManga, along with details regarding the annotation process.
    \item Section~\ref{supp_sec:evaluation} clarifies the character-clustering evaluation process for \textit{Magi} and the baselines.
    \item Section~\ref{supp_sec:ocr} provides information on the OCR model and the synthetic data used to train it.
\end{itemize}

\section{Panel ordering}
\label{supp_sec:panel_ordering}
As mentioned in Section~\ref{sec:transcript}, the idea behind our approach to order panels is to represent them as a directed acyclic graph (DAG). Each directed edge in this graph, represents the \textit{relative} reading order of the panels. In other words, if there is an edge from panel $i$ to panel $j$, then panel $i$ is read \textit{before} panel $j$ (not necessarily \textit{immediately} before, just before). Once such a graph is constructed, the reading order is simply given by the Topological Ordering~\cite{kahn1962topological} of this graph. To see why, consider the definition of Topological Ordering:
\begin{quote}
    A topological ordering of a directed graph is a linear ordering of its vertices such that for every directed edge $(i,j)$ from vertex $i$ to vertex $j$, $i$ comes before $j$ in the ordering.
\end{quote}
Therefore, once the topological ordering is computed for our DAG, all the panels $i$ that should be read before panel $j$, will [by definition] appear before $j$ in this ordering, thus giving us the overall reading order. The question, of course, is how to construct such a graph in the first place.

\subsection{Constructing the DAG}
\label{supp_sec:construct_dag}
Consider a directed graph $\mathcal{G}$, where each vertex corresponds to a panel on the page. To determine the edges of this graph, we need to determine the relative reading order between every pair of panels. This can be done by leveraging prior knowledge about manga layout. Specifically, the fact that the panels are read top to bottom, right to left. 

\subsubsection{Determining the edge between a pair of non-overlapping panels}
\label{supp_sec:nonoverlapping}
\begin{figure}[H]
    \centering
    \includegraphics[width=\linewidth]{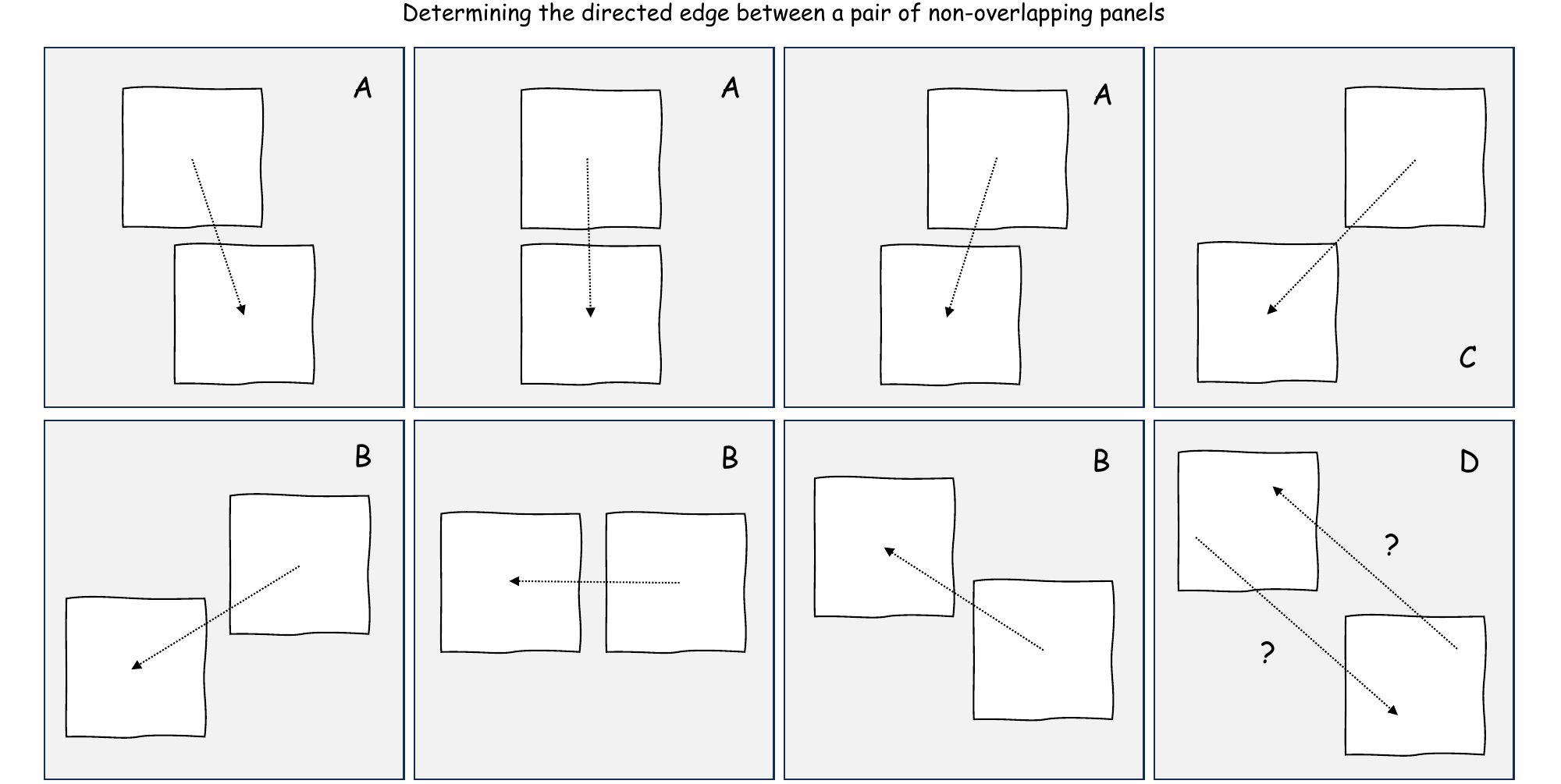}
    \caption{\textbf{Relative reading order between a pair of non-overlapping panels:} (A) If two panels are not strictly left or right of each other, but are strictly above or below, then the panel which is above is read before the panel which is below; (B) If the two panels are not strictly above or below, but are strictly left or right, then the panel which is on the right is read before the panel which is on the left; (C) If two panels are top-right and bottom-left of each other, then the top-right panel is always read before the bottom-left; (D) If two panels are top-left and bottom-right of each other, it is ambiguous (while one is above, the other is on the right and it is unknown as to which of the two gets priority, see Section~\ref{supp_sec:topleft}). 
    }
    \label{fig:directed_edges}
\end{figure}

Given two non-overlapping panels $i,j$ the edge $e_{i\rightarrow j}$ can be determined using the following rules in the given order of priority, where $e_{i\rightarrow j} = 1$ if $i$ is read before $j$ i.e.\ there is an edge from $i$ to $j$, and $e_{i\rightarrow j} = 0$ if $i$ is not read before $j$ i.e.\ there is no edge from $i$ to $j$. Figure~\ref{fig:directed_edges} provides supporting visualisations.

\begin{enumerate}
    \item If panel $i$ is strictly above panel $j$, and panel $j$ is not strictly to the right of $i$, $e_{i\rightarrow j} = 1$.
    \item If panel $j$ is strictly above panel $i$, and panel $i$ is not strictly to the right of $j$, $e_{i\rightarrow j} = 0$.
    \item If panel $i$ is strictly to the right of panel $j$, and panel $j$ is not strictly above $i$, $e_{i\rightarrow j} = 1$.
    \item If panel $j$ is strictly to the right of panel $i$, and panel $i$ is not strictly above $j$, $e_{i\rightarrow j} = 0$.
    \item If panel $i$ is strictly above panel $j$, \textit{but} panel $j$ is strictly to the right of $i$, see Section~\ref{supp_sec:topleft}.
    \item If panel $j$ is strictly above panel $i$, \textit{but} panel $i$ is strictly to the right of $j$, see Section~\ref{supp_sec:topleft}.
\end{enumerate}

\subsubsection{Determining the edge between a pair of overlapping panels}
\label{supp_sec:overlapping}
\begin{figure}[H]
    \centering
    \includegraphics[width=0.5\linewidth]{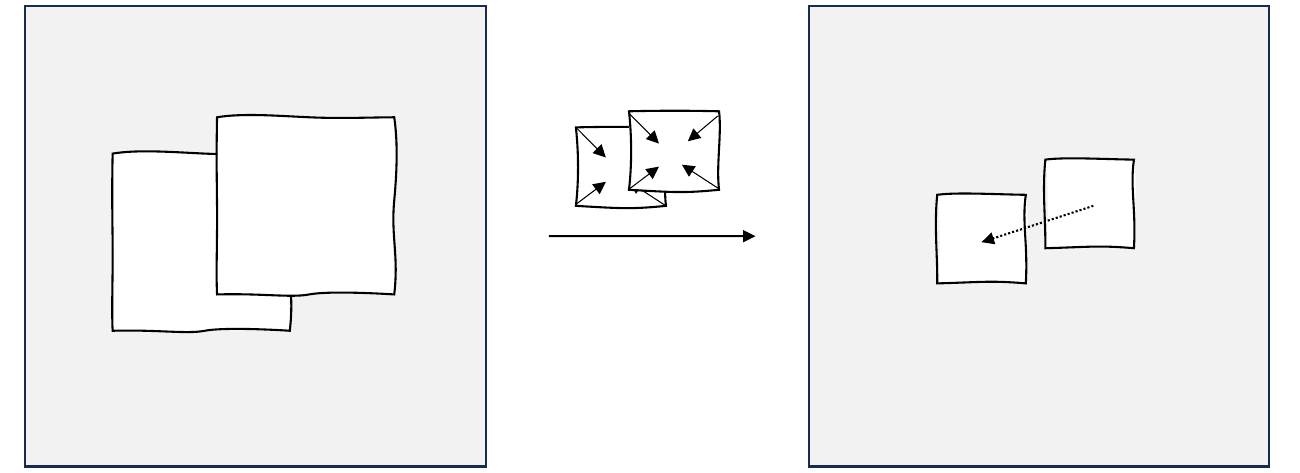}
    \caption{\textbf{Handling overlapping panels to determine the reading order:} An example of eroding overlapping panels until they don't intersect to faithfully determine the relative reading order.}
    \label{fig:erosion}
\end{figure}

In Section~\ref{supp_sec:nonoverlapping}, we determined the directed edge between two non-overlapping panels using ``strictly" constraints i.e. the panels must be \textit{strictly} above/below/left/right of one another. However, in the case of overlapping panels, these constraints cannot be applied. In order to determine the relative reading order of two overlapping panels, we propose to relax this ``strictly" constraint to ``largely" i.e.\ \textit{largely} above/below/left/right. The trick to implement this is very simple: If two panels intersect, apply morphological erosion~\cite{serra1982image} to their polygons until they do not intersect and then use the rules described in Section~\ref{supp_sec:nonoverlapping}. Figure~\ref{fig:erosion} shows an example. Note that this morphological erosion is only temporary and only applied while considering a pair of panels. After determining the relative order of these panels, their polygons \textit{must} be reset.

\subsubsection{Determining the edge between panels which are top-left and bottom-right of each other}
\label{supp_sec:topleft}
\begin{figure}[H]
    \centering
    \includegraphics[width=0.5\linewidth]{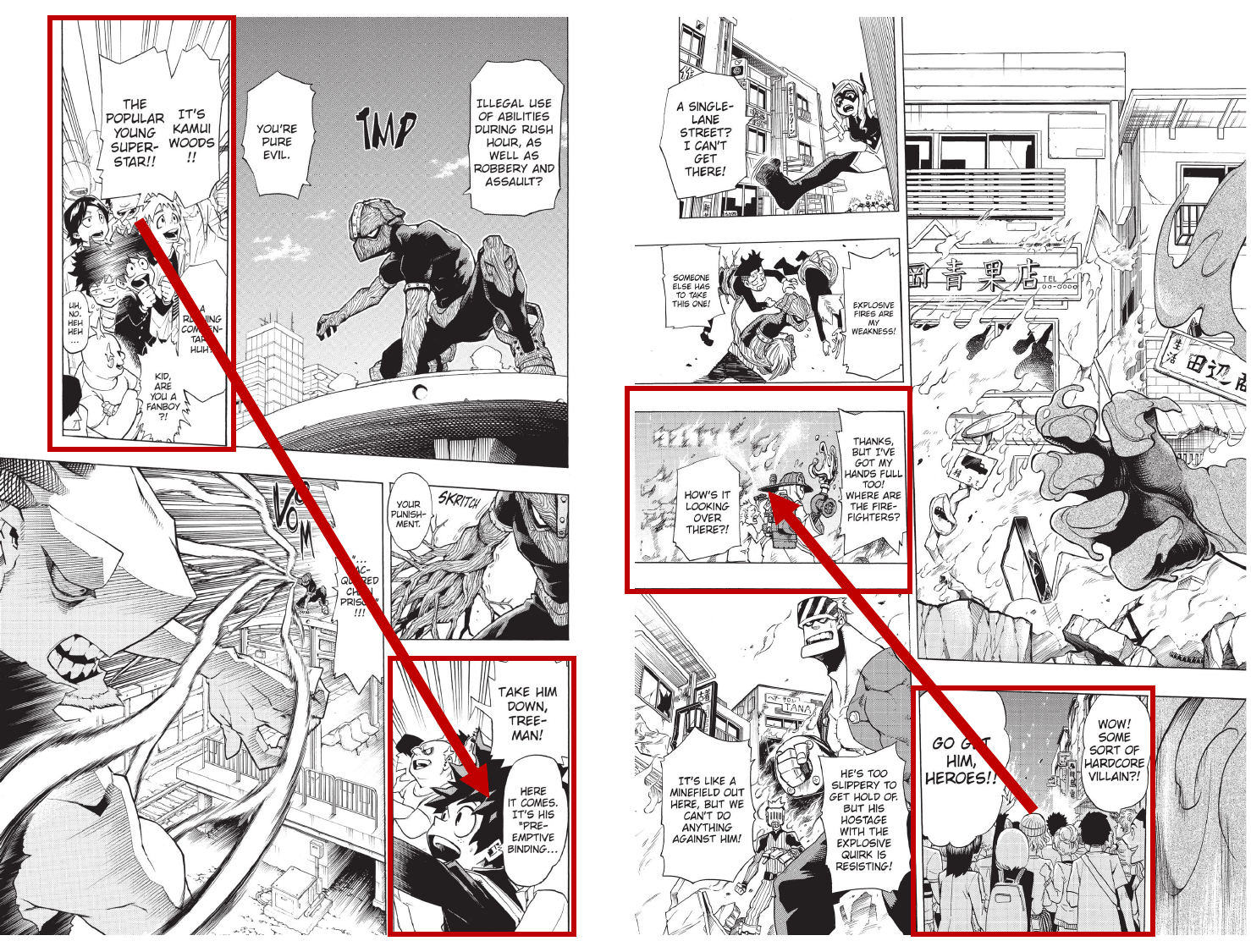}
    \caption{\textbf{Reading order when the panels are top-left and bottom-right:} A pair of panels that are top-left and bottom-right of each other can have a different relative order, depending on the layout of other panels. Specifically, the relative order here is evident from the partition of the \textit{entire} page. The left page is partitioned into top/bottom via horizontal whitespace, while the right page is partitioned into left/right via a vertical whitespace.}
    \label{fig:topleftbottomright}
\end{figure}

In the case where panel $i$ is above panel $j$, while panel $j$ is to the right of panel $i$, the relative reading order is ambiguous. Our prior knowledge tells us that top takes priority over bottom and right takes priority over left, however, it does not tell us what takes priority between top and right. Figure~\ref{fig:topleftbottomright} shows examples where the relative reading order between such panels can be $i\rightarrow j$ or $j\rightarrow i$ depending on the layout.

\noindent To disambiguate this special case, we need to partition the image via ``cuts". In detail, if panel $i$ is above and to the left of panel $j$, then,

\begin{enumerate}
    \item Try to ``cut" the image horizontally. If one or more cuts are possible, do any of them split $i$ and $j$ into two partitions (top and bottom)? If yes, then $e_{i\rightarrow j} = 1, e_{j\rightarrow i} = 0$.
    \item Try to ``cut" the image vertically. If one or more cuts are possible, do any of them split $i$ and $j$ into two partitions (right and left)? If yes, then $e_{i\rightarrow j} = 0, e_{j\rightarrow i} = 1$.
    \item If no cuts are possible, there is some overlap in the panels. Slightly erode all panels until a ``cut" is possible, and try again.
\end{enumerate}

\subsection{Examples of constructed DAGs and topological ordering}
Figure~\ref{fig:dag_and_ordering} shows some examples of the constructed DAGs using the approach described in Section~\ref{supp_sec:construct_dag}. We also show the computed topological ordering, which gives us the reading order of these panels.
\begin{figure}[H]
    \centering
    \includegraphics[width=\linewidth]{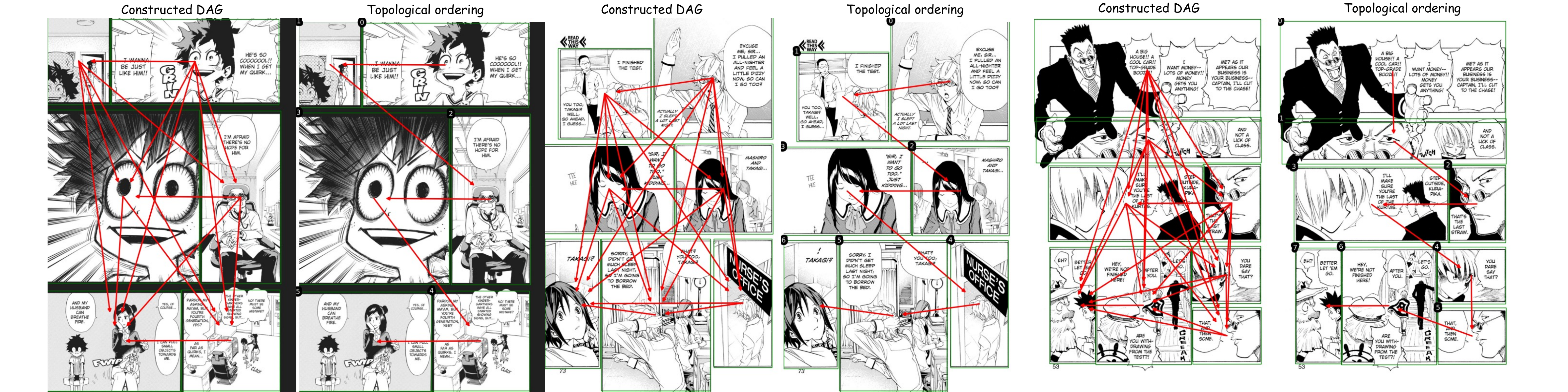}
    \caption{\textbf{DAGs and predicted reading order:} Examples of the constructed DAGs using the method described in Section~\ref{supp_sec:construct_dag} and the resulting topological/reading order. The panels shown are predictions from our model.}
    \label{fig:dag_and_ordering}
\end{figure}

\section{PopManga splits and annotation}
\label{supp_sec:popmanga}
In the following we catalogue the manga name and chapters used for the \textit{Dev}, \textit{Test-S}, \textit{Test-U} subsets of PopManga, as well as the annotation process.

\subsection{Data splits}

Table~\ref{tab:data_splits} contains a list of manga names and their chapters in PopManga. For \textit{Test-S} and \textit{Test-U} subsets we chose manga that are available on MangaPlus by Shueisha~\cite{MangaPlus}, an official source to read English manga globally, and we selected the first two chapters because these chapters are available to everyone for free.

Note that the \textit{Test-S} chapters are from the ``seen" manga i.e.\ the model has seen other chapters from this manga during training, whereas the \textit{Test-U} chapters are from the ``unseen" manga i.e.\ the model has never seen any chapter from this manga during training. In total there are over 2000 chapters in \textit{Dev}, 30 chapters in \textit{Test-S} and 20 chapters in \textit{Test-U}.

\begin{table}[H]
\centering
\resizebox{\columnwidth}{!}{%
\begin{tabular}{llll}
Manga                          & Dev                                                      & Test-S & Test-U \\ \hline
A Silent Voice                 & 1-38                                                       &     &      \\
Assassination Classroom        &                                                            &     & 1-2  \\
Bakuman                        & 3-13                                                       & 1-2 &      \\
Berserk                        & 0.01-0.16, 1-22                                            &     &      \\
Black Clover                   & 3-38                                                       & 1-2 &      \\
Bleach                         &                                                            &     & 1-2  \\
Blue Exorcist                  &                                                            &     & 1-2  \\
Blue Lock                      & 1-38                                                       &     &      \\
Blue Period                    & 1-12                                                       &     &      \\
Boku No Hero Academia          & 3-38                                                       & 1-2 &      \\
Bungo Stray Dogs               & 1-38                                                       &     &      \\
Chainsaw Man                   & 3-38                                                       & 1-2 &      \\
Chihayafuru                    & 1-38                                                       &     &      \\
Cross Game                     & 1-10                                                       &     &      \\
Dandadan                       & 1-32                                                       &     &      \\
Death Note                     & 3-38                                                       & 1-2 &      \\
Delicious In Dungeon           & 1-15                                                       &     &      \\
Detective Conan                & 1-38                                                       &     &      \\
Doraemon                       & 1-38                                                       &     &      \\
Dr Gray Man                    & 1-38                                                       &     &      \\
Dr Stone                       &                                                            &     & 1-2  \\
Dragonball                     & 3-38                                                       & 1-2 &      \\
Erased                         & 1-10                                                       &     &      \\
Eyeshield 21                   & 1-5                                                        &     &      \\
Food Wars                      &                                                            &     & 1-2  \\
Fullmetal Alchemist            & 1-38                                                       &     &      \\
Gekkan Shoujo Nozaki Kun       & 1-38                                                       &     &      \\
Given                          & 1-28                                                       &     &      \\
Grand Blue                     & 1-23                                                       &     &      \\
Haikyuu                        & 1-38                                                       &     &      \\
Hajime No Ippo                 & 1-38                                                       &     &      \\
\end{tabular}
\begin{tabular}{llll}
Manga                          & Dev                                                      & Test-S & Test-U \\ \hline
Hinamatsuri                    & 1-21                                                       &     &      \\
Horimiya                       & 1-7                                                        &     &      \\
Houseki No Kuni                & 1-38                                                       &     &      \\
Hunter X Hunter                & 3-38                                                       & 1-2 &      \\
Jigokuraku                     & 3-38                                                       & 1-2 &      \\
Jojo's Bizarre Part 1          & 3-38                                                       & 1-2 &      \\
Jojo's Bizarre Part 2          & 1-38                                                       &     &      \\
Jojo's Bizarre Part 3          & 1-37                                                       &     &      \\
Jojo's Bizarre Part 4          & 1-37                                                       &     &      \\
Jojo's Bizarre Part 5          & 1-37                                                       &     &      \\
Jojo's Bizarre Part 6          & 1-37                                                       &     &      \\
Jojo's Bizarre Part 7          & 1-37                                                       &     &      \\
Jojo's Bizarre Part 8          & 1-37                                                       &     &      \\
Jujutsu Kaisen                 & 3-37                                                       & 1-2 &      \\
Kagurabachi                    &                                                            &     & 1-2  \\
Kaguyasama Wa Kokurasetai      & 1-6                                                        &     &      \\
Kaichou Wa Maid Sama           & 3-37                                                       &     &      \\
Kamisama Hajimemashita         & 1-33,37,40,59,74                                           &     &      \\
Kimetsu No Yaiba               & 3-37                                                       & 1-2 &      \\
Kingdom                        & 1-37                                                       &     &      \\
Konjiki No Gash                & 1-37                                                       &     &      \\
Kuroko No Basket               &                                                            &     & 1-2  \\
Kuroshitsuji                   & 1-37                                                       &     &      \\
Made In Abyss                  & 1-37                                                       &     &      \\
Magi                           & 1-37                                                       &     &      \\
Mairimashita Iruma Kun         & 1-25                                                       &     &      \\
Naruto                         & 3-37                                                       & 1-2 &      \\
Nisekoi                        &                                                            &     & 1-2  \\
Noragami                       & 1-23,28-38,43                                              &     &      \\
One Piece                      & 3-37                                                       & 1-2 &      \\
\end{tabular}
\begin{tabular}{llll}
Manga                          & Dev                                                      & Test-S & Test-U \\ \hline
One Punch Man                  & 
\begin{tabular}[l]{@{}l@{}}
1-16,18-19,\\
22,24-25,\\
27-31,33,37,\\
39,42-43,46,\\
48-49,51,54-55
\end
{tabular}

&     &      \\
Oshi No Ko                     &                                                            &     & 1-2  \\
Pandora Hearts                 & 1-37                                                       &     &      \\
Parasyte                       & 1-37                                                       &     &      \\
Relife                         & 1-37                                                       &     &      \\
Shadows House                  & 1-37                                                       &     &      \\
Shingeki No Kyojin             & 1-37                                                       &     &      \\
Silver Spoon                   & 1-11                                                       &     &      \\
Skip Beat                      & 1-37                                                       &     &      \\
Slam Dunk                      & 1-26,45-53                                                 &     &      \\
Spy X Family                   & 3-37                                                       & 1-2 &      \\
Tensei Shitara Slime Datta Ken & 1-24                                                       &     &      \\
The World God Only Knows       & 1-37                                                       &     &      \\
To Your Eternity               & 1-37                                                       &     &      \\
Tokyo Ghoul                    & 3-37                                                       & 1-2 &      \\
Tokyo Revengers                & 1-37                                                       &     &      \\
Tsubasa Reservoir Chronicles   & 1-37                                                       &     &      \\
Vagabond                       & 1-10                                                       &     &      \\
Vinland Saga                   & 1-37                                                       &     &      \\
Witch Hat Atelier              & 1-37                                                       &     &      \\
World Trigger                  &                                                            &     & 1-2  \\
Xxxholic                       & 1-20                                                       &     &     
\end{tabular}%
}
\caption{\textbf{PopManga splits.} The numbers in the columns are the chapter numbers. For instance, Chapters 1-2 of \textit{Bakuman} are in the \textit{Test-S} set while Chapters 3-13 are in the \textit{Dev} set.}
\label{tab:data_splits}
\end{table}

\subsection{Annotation instructions}
\label{supp_sec:instructions}
For each page in PopManga, we annotate the following: (i) text boxes, (ii) character boxes, (iii) character-character positive pairs, (iv) text-character positive pairs. To label these pages we employed human annotators. They were provided with a modified version of the LISA annotation tool~\cite{Dutta19a}, as shown in Section~\ref{supp_sec:lisa}, and the following instructions:

\begin{enumerate}
    \item Before labelling a page, check if it is a ``story page". If it is not, ignore it. Examples of non-story pages include title pages, cover pages, table of content pages, bonus art pages etc. Basically, anything that is not a typical manga page and does not add to the story, ignore it.
    \item For each text ``block" put a bounding box around it. Note that the ``blocks" are not individual letters, nor individual lines, but rather anything that is spatially close-by and should be read together as a single statement. Within the block, we should be able to read the text from top-left to bottom-right, as is standard in English. Text is also not limited to just dialogues but also includes sound effects, scene texts etc.\ (although this distinction is not recorded).
    \item  For each character, put a bounding box around it. A character is not limited to humans only. Also annotate monsters, animals, or more broadly, anything that appears to be \textit{sentient}. Furthermore, draw a bounding box around characters even if they are only partially visible (e.g.\ only a hand is visible).
    \item For character-character matches, go over each panel in the reading order. For each character box (node) in the current panel, connect it (create an edge) to one of the character boxes in the previous panels, if they are the same character. This results in a set of connected components, where each connected component corresponds to a unique character identity.
    \item For text-character matches, \textit{read} the story in order, and connect text boxes to the box of the speaker.
\end{enumerate}

\subsection{Annotation process}
\label{supp_sec:annotation_process}
Given the instructions described in Section~\ref{supp_sec:instructions}, we were able to obtain high quality annotations for PopManga (subject to slight human error). The data was split among 13 different annotators, where each page had a main annotator and a reviewer.
The task of text-character matching was the most difficult of all for our annotators, due to unfamiliarity with manga and language barrier (English not being their first language). Due to this we were only able to annotate a small percentage of the total number of images for this task.

\subsection{Annotation tool}
\label{supp_sec:lisa}
To label bounding boxes, character-character association and text-character associations in PopManga, we modify the LISA annotation tool~\cite{Dutta19a} which already supports adding bounding boxes to the images. Specifically, we add functionality to simplify the \textit{association} annotation process (clicking two boxes adds an ``edge"), along with cosmetic changes to make it easier to use including colour coding for different classes of boxes, highlighting the boxes which belong to the same character when hovered, displaying text-character associations via lines, toggling on/off certain types of annotations to declutter the image. Then, we pack each manga chapter into a single LISA project. This allows the annotators to sequentially annotate the pages of the same chapter (by scrolling vertically) and do not need to context switch across different series.

\begin{figure}[H]
    \centering
    \includegraphics[width=0.9\linewidth]{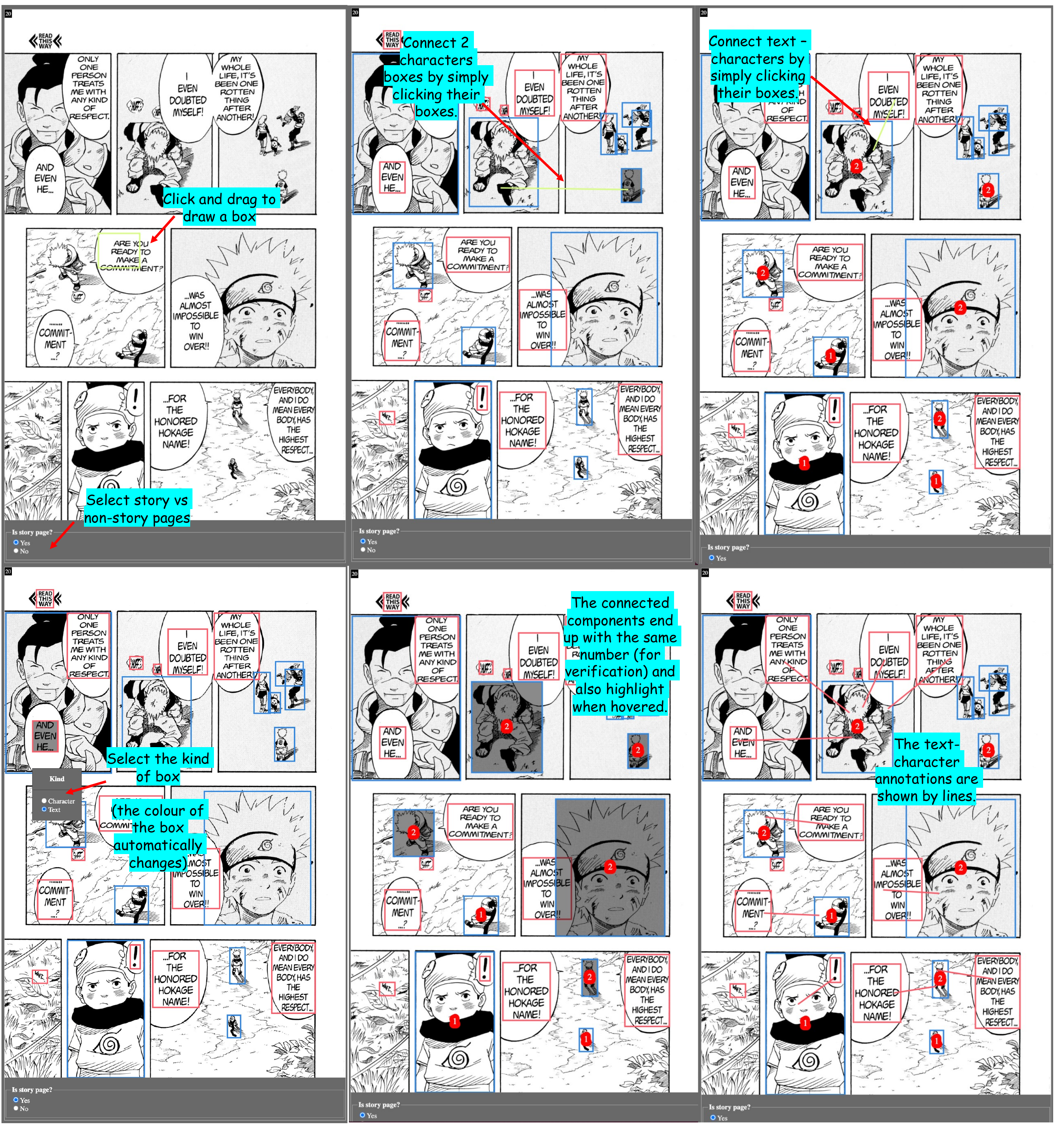}
    \caption{\textit{Annotation tool:} Modified version of the LISA annotation tool.}
    \label{fig:dag_and_ordering}
\end{figure}

\subsection{Comparison of PopManga annotations with Manga109}
\begin{figure}[H]
    \centering
    \includegraphics[width=0.8\linewidth]{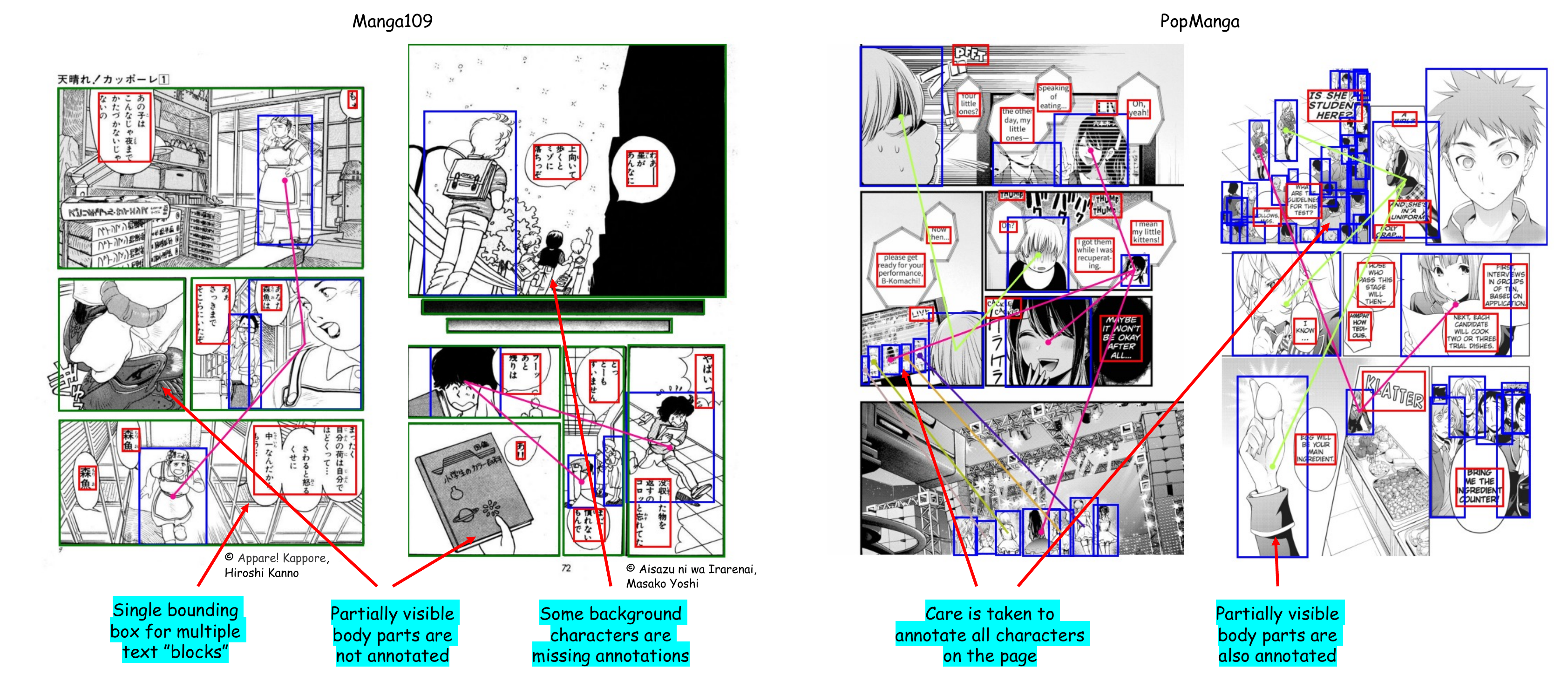}
    \caption{\textbf{Manga109 vs PopManga annotations:} We compare the annotations of Manga109 with PopManga and contrast the design decisions. The images in Manga109}
    \label{fig:manga109_vs_popmanga}
\end{figure}

In Figure~\ref{fig:manga109_vs_popmanga} we compare our design decisions for annotating PopManga with Manga109~\cite{manga109}. The main differences are: 
\begin{enumerate}
    \item In Manga109, boxes for texts span the entire multi-part speech balloon, whereas in ours, each box only covers a single text ``block". This discrepancy is due to the \textit{tategaki} writing system in Japanese, where characters/letters within a speech balloon are read top-bottom, right-left (just like the manga panels). Therefore, splitting a multi-part speech balloon into individual ``blocks" in Manga109 makes no difference to the reading order of individual characters/letters. However, in the translated English versions, the text within each ``block" is read left-right, top-bottom, but the blocks themselves follow top-bottom, right-left structure. It is therefore essential to split them into individual ``blocks" to properly perform OCR and generate the transcript.
    \item In PopManga, we made the decision to annotate everything that may be considered a character. This includes characters that are only partially visible, or background characters.
    \item In Manga109, the characters are given a globally unique ID, i.e.\ characters may be matched across several pages if desired. To reduce the annotation cost and the extra overhead to assign global IDs, we decided not to do this. Therefore, our character boxes only have a local ID (per page).
    \item Manga109 has additional annotations including face boxes, panels, onomatopoeia etc. which PopManga does not.
\end{enumerate}

\section{Some notes on clustering evaluation}
\label{supp_sec:evaluation}
Here we provide some clarifications regarding the evaluation process for character clustering. 
\subsection{Comparing Magi with baselines}
\begin{enumerate}
    \item All the baseline results presented are computed by using the model weights that have been open sourced by the prior works.
    \item Since prior works for character clustering only operate on crops of characters, we crop the original image using the ground truth character bounding boxes and then compute the per-crop embeddings, which are then used for evaluation.
    \item Since \textit{Magi} operates on the entire image and not crops, we first match the predicted character boxes to ground truth via Hungarian matching, and then use the features corresponding to the matched predicted characters for evaluation. This is also how we are able to compute results on Manga109 (face), even though our method is not trained to explicitly predict faces.
\end{enumerate}

\subsection{Implementation of metrics}
\begin{enumerate}
    \item We adopt the implementation of each metric from Pytorch Metric Learning~\cite{Musgrave2020PyTorchML}. The metrics are computed per page and averaged across the entire test set.
    \item While computing AMI and NMI, we do not assume prior knowledge regarding the number of unique identities present on the page. This is because we wish to evaluate the model's ability to form decisive clusters without knowing the number of clusters apriori. To facilitate this, we compute a character-character similarity matrix for each page and threshold this matrix to give us binary values for whether two characters are the same or not (thus, giving us clusters). The AMI and NMI results reported in Table~\ref{tab:char_clus} for baselines are by picking the best threshold for each method for each test set (where ``best" corresponds to the threshold value with maximum AMI results). The threshold for \textit{Magi} is always $0.65$.
\end{enumerate}

\section{OCR}
\label{supp_sec:ocr}
\begin{figure}[H]
    \centering
    \includegraphics[width=\linewidth]{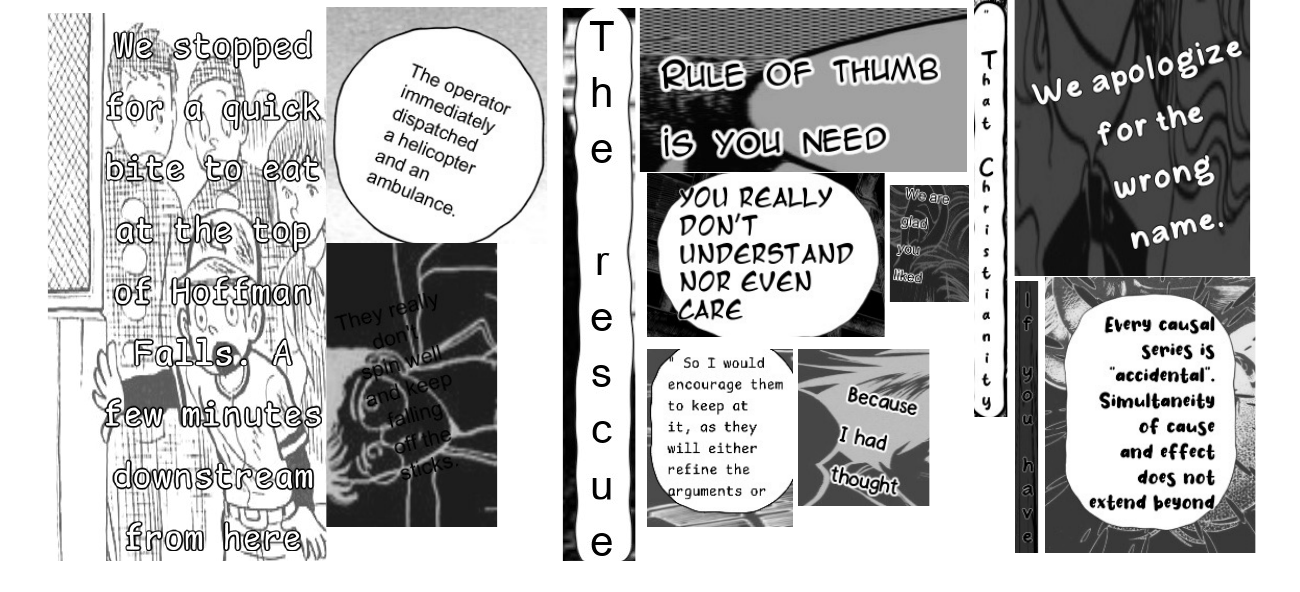}
    \caption{\textbf{Synthetic data for OCR:} Examples of generated synthetic images. The images are generated at varying resolutions and aspect ratios to make the model robust.}
    \label{fig:ocr}
\end{figure}

We adapt the synthetic data generation pipeline in~\cite{manga_ocr} to generate $\approx$1M synthetic images to finetune the TrOCR~\cite{li2021trocr} model. We show some examples in Figure~\ref{fig:ocr}.

\end{document}